\documentclass{article} % For LaTeX2e
\usepackage{iclr2025_conference,times}

\usepackage{amsmath,amsfonts,bm}

\def\eqref#1{equation~\ref{#1}}
\def\1{\bm{1}}

\DeclareMathAlphabet{\mathsfit}{\encodingdefault}{\sfdefault}{m}{sl}
\SetMathAlphabet{\mathsfit}{bold}{\encodingdefault}{\sfdefault}{bx}{n}

\usepackage{hyperref}
\usepackage{url}
\usepackage{graphicx}
\usepackage{amsmath}
\usepackage{amssymb}
\usepackage[font=normalsize]{caption}
\usepackage{textcomp}

\usepackage{subcaption}
\usepackage{xcolor}
\usepackage{algorithm}
\usepackage{algpseudocode}
\usepackage{booktabs}
\usepackage{makecell}
\usepackage{colortbl}
\definecolor{Gray}{gray}{0.9}
\definecolor{LightCyan}{rgb}{0.75,1,1}
\newcolumntype{y}{>{\columncolor{LightCyan}}c}

\usepackage[many]{tcolorbox}  
\newtcolorbox{boxL}{
    fontupper = \color{black},
    rounded corners,
    arc = 6pt,
    colframe = black!50, 
    boxrule = 0pt, 
    bottomrule = 4.5pt ,
    breakable,
}

\title{Steering Large Language Models between Code Execution and Textual Reasoning}

\author{Yongchao Chen \\
  MIT / Harvard \\
  \small\texttt{yongchaochen@fas.harvard.edu}
  \And
  Harsh Jhamtani \\
  Microsoft \\
  \small\texttt{hjhamtani@microsoft.com} 
  \And
  Srinagesh Sharma \\
  Microsoft \\
  \small\texttt{srsharm@microsoft.com}
  \And
  Chuchu Fan \\
  MIT \\
  \small\texttt{chuchu@mit.edu}
  \\\And
  Chi Wang \\
  Google DeepMind \\
  \small\texttt{chi@chiwang.cc} \\  }

\iclrfinalcopy % Uncomment for camera-ready version, but NOT for submission.
\begin{document}

\maketitle

% Create the main Table of Contents (empty)
% but suppress adding contents for the main article
\addtocontents{toc}{\protect\setcounter{tocdepth}{-1}}

\begin{abstract}
While a lot of recent research focuses on enhancing the textual reasoning capabilities of Large Language Models (LLMs) by optimizing the multi-agent framework or reasoning chains, several benchmark tasks can be solved with 100\% success through direct coding, which is more scalable and avoids the computational overhead associated with textual iterating and searching. Textual reasoning has inherent limitations in solving tasks with challenges in math, logics, optimization, and searching, which is unlikely to be solved by simply scaling up the model and data size. The recently released OpenAI GPT Code Interpreter and multi-agent frameworks such as AutoGen have demonstrated remarkable proficiency of integrating code generation and execution to solve complex tasks using LLMs. However, based on our experiments on 7 existing popular methods for steering code/text generation in both single- and multi-turn settings with 14 tasks and 6 types of LLMs (including the new O1-preview), currently there is no optimal method to correctly steer LLMs to write code when needed. We discover some interesting patterns on when models use code vs. textual reasoning with the evolution to task complexity and model sizes, which even result in an astonishingly inverse scaling behavior. We also discover that results from LLM written code are not always better than using textual reasoning, even if the task could be solved through code. To mitigate the above issues, we propose three methods to better steer LLM code/text generation and achieve a notable improvement. The costs of token lengths and runtime are thoroughly discussed for all the methods. We believe the problem of steering LLM code/text generation is critical for future research and has much space for further improvement. Project Page, Datasets, and Codes are available at \url{https://yongchao98.github.io/CodeSteer/}.
\end{abstract}

\begin{figure*}[h]
  \centering
  \includegraphics[width=0.95\linewidth]{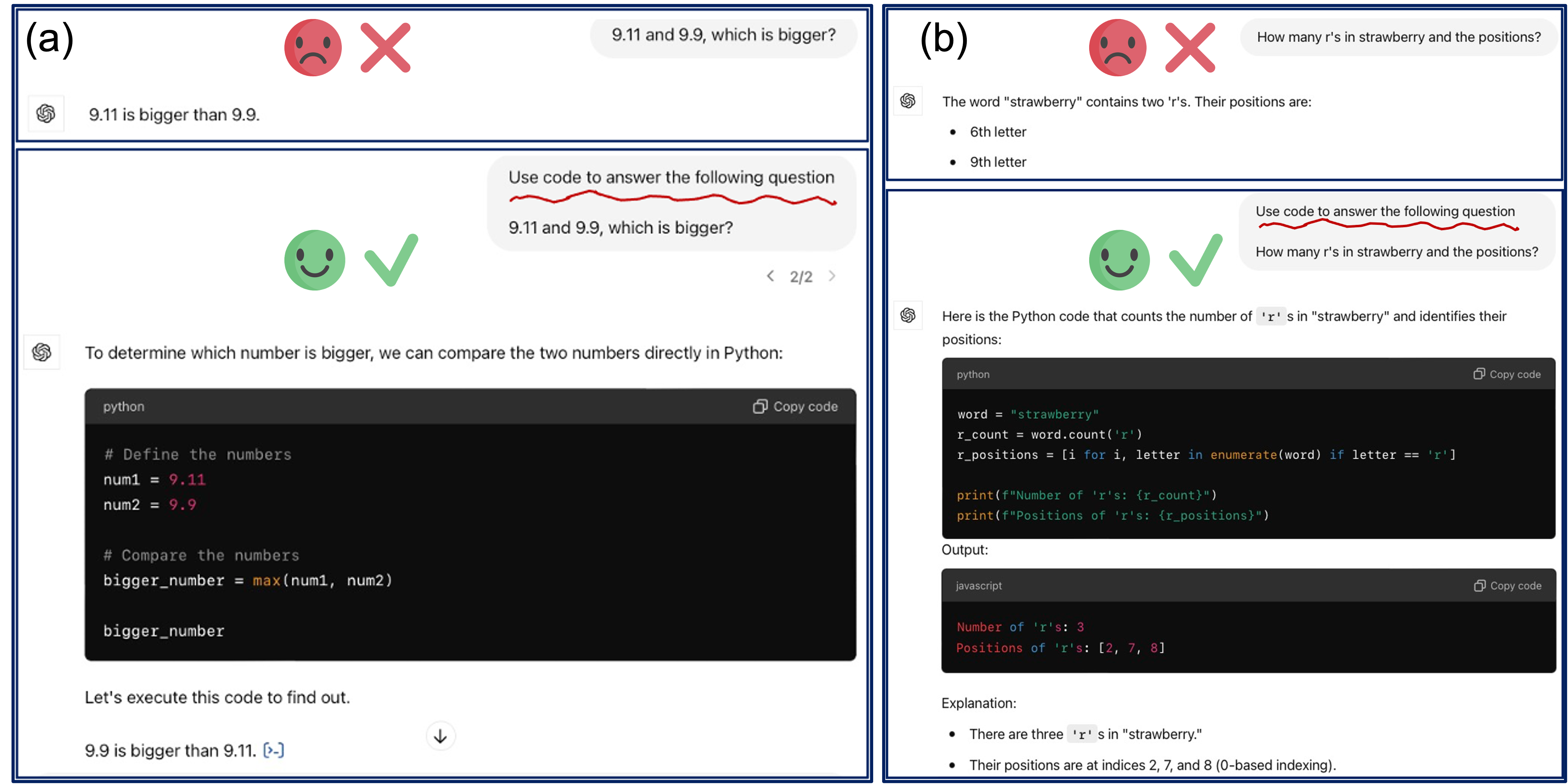}
   \caption{The cases that GPT-4o makes simple mistakes by direct textual reasoning but can reliably solve the problem with prompted to use code.}
   \label{fig:GPT4o-makes-simple-mistakes-in-number-letter}
\end{figure*}

\section{Introduction}
The rapid progress of LLMs has inspired a great quantity of research in building general language-guided agents that can solve various tasks automatically~\citep{autogen,chain-of-code,Tree-of-thought, graph-of-thoughts}. While the capabilities of these LLM-based agents have been largely enhanced by tuning the agent frameworks~\citep{CogEval,mixture-of-agents}, reasoning chains~\citep{chain-of-code,Tree-of-thought,graph-of-thoughts}, visual and spatial abilities~\citep{VLM-1,VLM-2}, and input prompts~\citep{meta-prompting,PROMST,black-box-prompt-optimize}, the best LLMs still make mistakes on simple tasks~\citep{nature-paper-llm-not-reliable}, such as the recently popular topics of '9.11' and '9.9' numerical comparison and 'r' letter count in 'strawberry', as shown in Fig~\ref{fig:GPT4o-makes-simple-mistakes-in-number-letter}. However, after reviewing all the tested tasks from previous papers, we detect that nearly half of the tasks can be completely solved by coding, such as Blocksworld~\citep{planbench}, Game 24~\citep{LATS}, and Logical Deduction~\citep{big-bench-hard}.

Text is suitable for semantic analysis and commonsense reasoning, but is not the best format for precise computation and planning, symbolic manipulation, and algorithmic processing and optimization~\citep{llm-cannot-plan-1,autotamp}. Conversely, programs excel in rigorous operations, and can outsource intricate calculations to specialized tools like equation solvers. Since recent LLMs are well trained at code generation~\citep{codeplan-code-use-llm}, one question that comes up is whether querying LLMs to generate code can be more effective than textual reasoning. 

In this study, we emphasize that textual reasoning has inherent limitations in solving tasks that involve math, logic, and optimization, where coding can often provide a better solution. For example, Fig~\ref{fig:GPT4o-makes-simple-mistakes-in-number-letter} presents two typical examples that the ChatGPT of GPT-4o makes mistakes by direct textual reasoning but easily solves the problem after prompted to use code. Recent studies also show that using a code-based framework enhances LLMs' logical reasoning performance, even in commonsense reasoning tasks~\citep{llm+code=commense-learner,code-as-policies,Program-of-thoughts-prompting}. Targeting text as the only output modality has limitations in the scalability of task complexity. As shown in Appendix~Fig~\ref{fig:number_multiply_varied_digit}, even if the model of OpenAI series scales from GPT-3.5 to O1, LLM still easily fails once the task complexity grows. In these tasks, coding is a scalable and complete solution.

Guiding LLMs to choose between code generation/execution and textual reasoning remains a challenging problem, as common questions lack prior cues for either approach. Recent OpenAI GPT models address this issue by augmenting the Code Interpreter (CI) function, where models are trained to use an integrated coding platform as part of their reasoning~\citep{gpt-4,llama-3-report}. Once the model generates code, the platform executes it and returns the results for further processing. This iterative process of generating code and text continues until the final answer is reached. In addition to GPT CI, multi-agent frameworks like AutoGen~\citep{autogen} query the LLM itself, using a specific system prompt to decide when to generate code, which is then executed based on predefined rules. More related work are specifically discussed in Appendix Section~\ref{appendix sec: More Related Work}.

In this paper, we perform an in-depth investigation into the effectiveness of LLMs in steering between use of textual reasoning and code generation/execution across 14 diverse tasks requiring mathematical, verbal, and planning capabilities, using 6 types of LLMs (O1-preview, GPT-4o~\citep{gpt-4}, GPT-4o-mini, GPT-3.5~\citep{gpt-3}, Claude-sonnet~\citep{claude}, Mixtral-8x7b~\citep{mixtral}). The key contributions and findings of our work are:

1. \textbf{Existing methods struggle to optimally decide when to write code or use textual reasoning}: 
We evaluated 10 different methods to steer LLMs to use code when required by conducting experiments across 14 datasets/tasks and 6 LLMs (with and without built-in CI, prompt modifications to favor code over text, multi-turn code refinement, etc.). Our experiments reveal that there is no single best method across the board. Additionally, we also analyze the trade-offs in token length and runtime against accuracy for each method, which can serve as a guidance to pick the method of choice based on budget and performance expectations. 
% Our experiments suggest that there is no universally optimal method for steering LLMs to generate code when necessary. Our experiments across 14 datasets/tasks, 10 methods, and 6 models (with and without built-in CI, prompt modifications to favor code over text, multi-turn code refinement, etc.) reveal that each method performs differently depending on the setup. We also analyze the trade-offs in token length and runtime for each method.

2. \textbf{Forcing LLMs to write code is not guaranteed to give more accurate results compared to textual reasoning:} Our experiments suggest that prompting LLM to answer directly with code can sometimes lead to worse overall accuracy. We discuss several contributing reasons for such behavior. Firstly, writing correct code is tough in certain tasks, such as robot task planning. Secondly, code format can limit the space of output tokens, potentially hindering the reasoning ability of LLMs ~\citep{speak-freely}. 
Moreover, we observe that LLMs sometimes generate code that resembles more of textual reasoning rather than containing any functional implementations.
% Finally, LLMs sometimes will not follow code writing prompts and still answer with text, though it is rare.

3. \textbf{Our experiments reveal the patterns on when LLMs use code vs. textual reasoning as a function of factors such as task complexity and model sizes:} Surprisingly, when augmented with CI, smaller models like GPT-3.5 sometimes outperform larger ones like GPT-4o, illustrating an inverse scaling behavior, contrary to previous studies~\citep{scaling-law-1}. This phenomenon appears to be linked to the varied LLM's confidence in its textual reasoning ability. As a result, GPT-3.5 outperforms GPT-4o in Game 24 and Number Multiplying tasks (inverse scaling behavior).

4. \textbf{Mixing code and textual reasoning, and multi-turn refinement:} Inspired by previous work that utilize multi-agent framework to refine answers~\citep{chen2023reconcile,mixture-of-agents}, we propose optimized methods like assembling coding and textual reasoning together resulting into improvements across 6 models. We also show that multi-turn execution/refinement~\citep{autotamp,llm-can-help-plan-in-framework} improves the performance. However, we also find that these methods are limited by the LLM's inherent capabilities and the frequency of code usage, which in turn increases runtime costs and token lengths.

\section{OpenAI GPT Code Interpreter struggles in code/text choices}
Are current methods capable of effectively switching between code execution and textual reasoning? Since OpenAI GPT CI is the current most popular and effective method for steering code/text generation~\citep{mathcoder,code-based-self-verify}, we carry out a detailed exploration on its characteristics and limitations. In this section, we use Number Multiplying (calculating number multiplication) and Game 24 (outputting an equation that evaluates to 24 with the given set of integers) as representative tasks because they are simple to describe to LLMs without being affected by different prompt types, and their complexity can be easily adjusted. The question prompt is the same as the original dataset~\citep{LATS,Tree-of-thought} without the extra hints for code/text steering.

\begin{figure*}[h]
  \centering
   \includegraphics[width=0.8\linewidth]{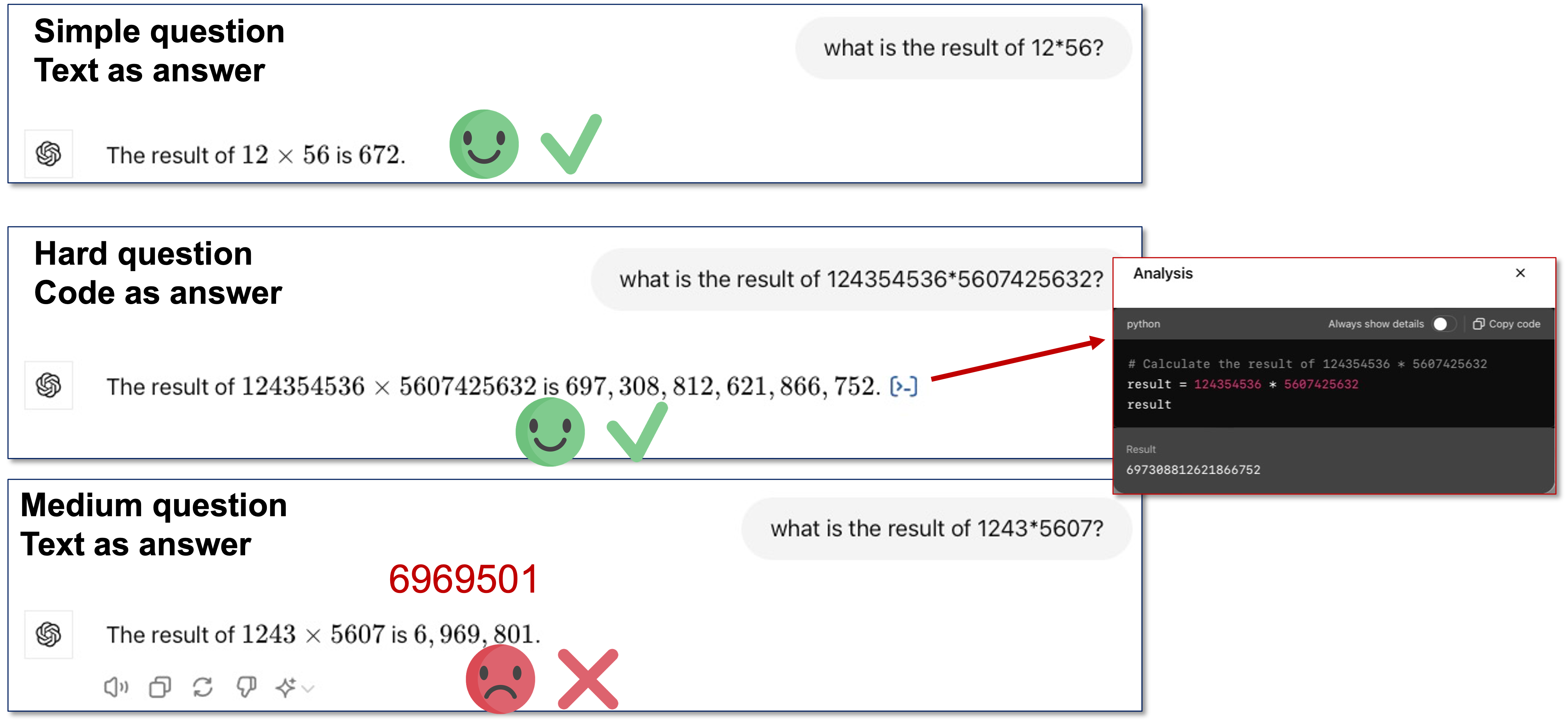}
   \caption{GPT-4o CI tends to handle simple Number Multiplying tasks with text and complex tasks with code. However, it often fails with medium-difficulty questions, where it is overconfident and chooses not to use code when needed.}
   \label{fig:Evolution-with-complexity-number-multiply}
\end{figure*}

\begin{figure*}[h]
  \centering
   \includegraphics[width=0.9\linewidth]{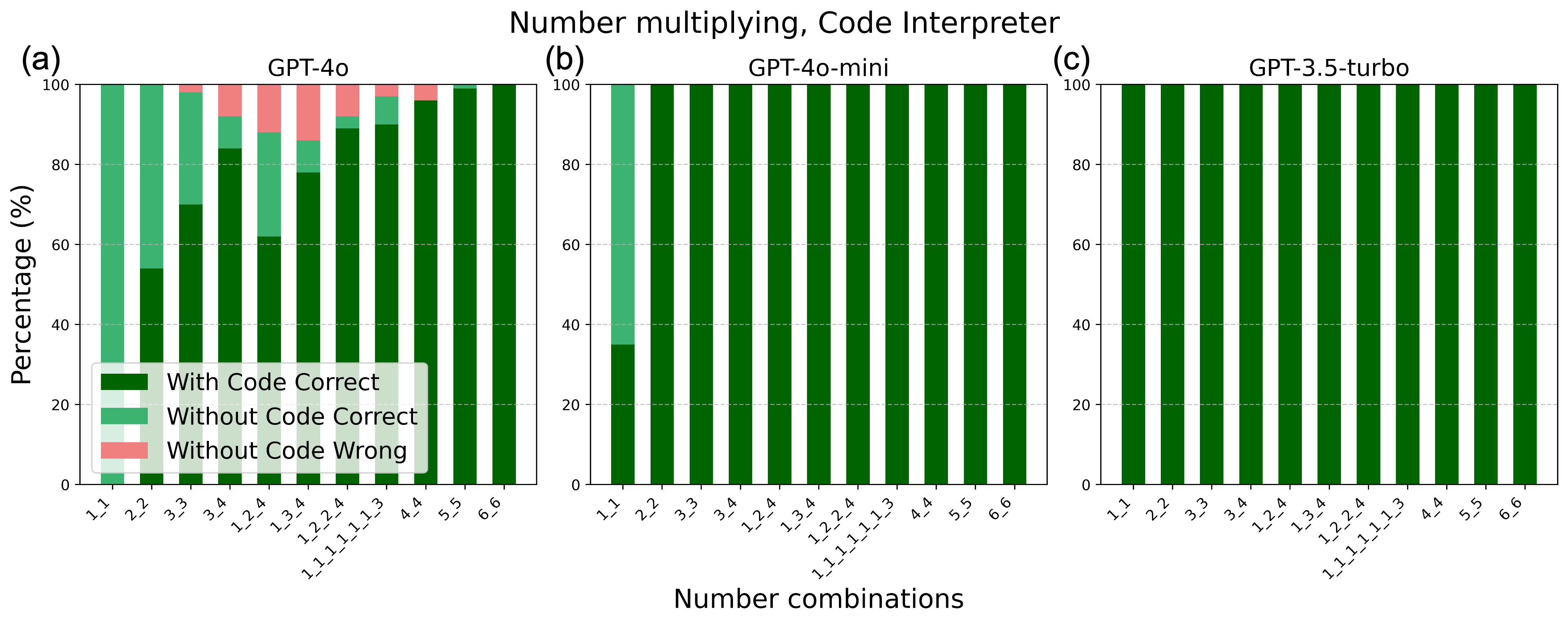}
   \caption{Success rates and code usage rates of GPT CI in Number Multiplying task across varied task complexity. The labels on the x-axis represent the number of digits in the numbers being multiplied. For example, '3\_4' means a three-digit number multiplied by a four-digit number. From the left to the right of x-axis, the digit numbers of multiplied values increase, representing increasing task complexity. The success and failure cases are visualized with green and red colors, respectively.}
   \label{fig:number_multiply_code_inter}
\end{figure*}

\begin{figure*}[h]
  \centering
   \includegraphics[width=0.75\linewidth]{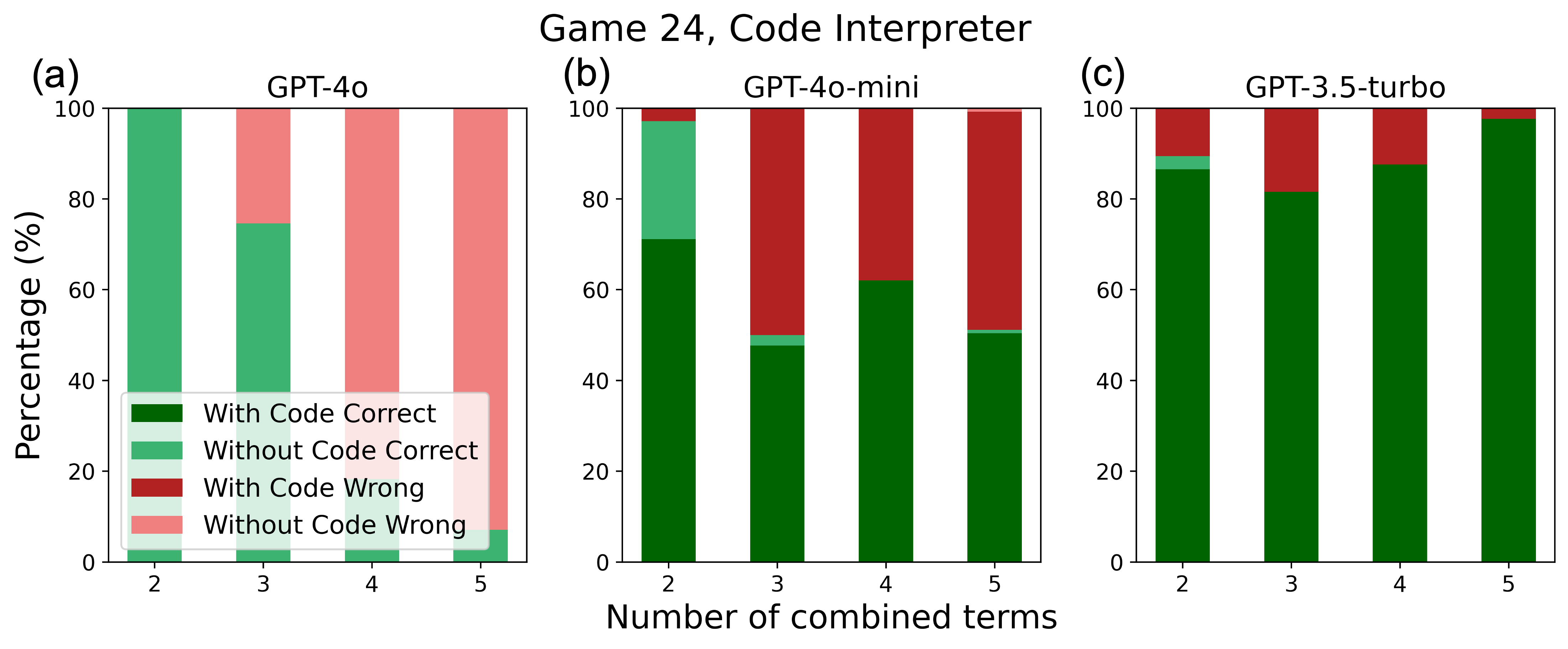}
   \caption{Success rates and code usage rates of OpenAI CI in Game 24 task across varied task complexity. The labels on the x-axis represent the number of terms/values to be combined to form value 24, presenting an increasing task complexity from the left to the right side. The success and failure cases are visualized with green and red colors, respectively.}
   \label{fig:game24_code_inter}
\end{figure*}

\subsection{Evolution with task complexity}
\label{sec:Evolution with task complexity}
We observe an intriguing property of GPT CI: its decision to use code depends on the complexity of the task, as shown in Fig~\ref{fig:Evolution-with-complexity-number-multiply}. GPT-4o CI chooses to handle simple Number Multiplying questions with text and complex questions with code, resulting in correct answers. However, it fails in medium-difficulty questions since it tends to be overconfident and chooses to answer the question via textual reasoning, which sometimes is wrong. Fig~\ref{fig:number_multiply_code_inter}a visualizes this phenomenon quantitatively. We adjust the task complexity via changing the number of digits in the numbers being multiplied. GPT-4o answers all multiplication questions correctly using text for small numbers (left side) and generates code for very large numbers (right side). Errors occur when the numbers are neither too large nor too small, causing GPT-4o to struggle with deciding whether to use code. In the Number Multiplying task, generating code is straightforward, leading to correct answers when code is used. However, GPT-4o's tendency to be overconfident and avoid using code in medium-difficulty questions results in occasional failures.

\subsection{Inverse scaling behavior of model sizes}
\label{sec:Inverse scaling law of model sizes}
Based on the above property, we also see that in some tasks smaller models outperform larger models when all augmented with CI, which is inverse to the well-known scaling law in LLMs~\citep{scaling-law-1}. In Fig~\ref{fig:number_multiply_code_inter}, both GPT-4o-mini and GPT-3.5 achieve 100\% success rates across all the task complexity. Compared to GPT-4o, GPT-4o-mini and GPT-3.5 are more conservative so that they generate code all the time when encountering slightly complex questions. This phenomenon also appears in Game 24 task (Fig~\ref{fig:game24_code_inter}) where GPT-4o generates text in all the questions while GPT-4o-mini and GPT-3.5 generate code in most cases. In Game 24, coding is also a more reliable method than textual reasoning. Hence, GPT-4o-mini and GPT-3.5 notably outperform GPT-4o, especially when task complexity is high. In the Game 24 task, both GPT-4o-mini and GPT-3.5 tend to generate code more frequently as task complexity increases, further supporting the conclusion in Section~\ref{sec:Evolution with task complexity}. We hypothesize that GPT-4o will increasingly use code as task complexity rises. However, our additional tests (not included in the paper) indicate that GPT-4o only begins using code to solve Game 24 problems when the complexity of the tasks becomes exceedingly high.

\begin{figure*}[h]
  \centering
   \includegraphics[width=0.85\linewidth]{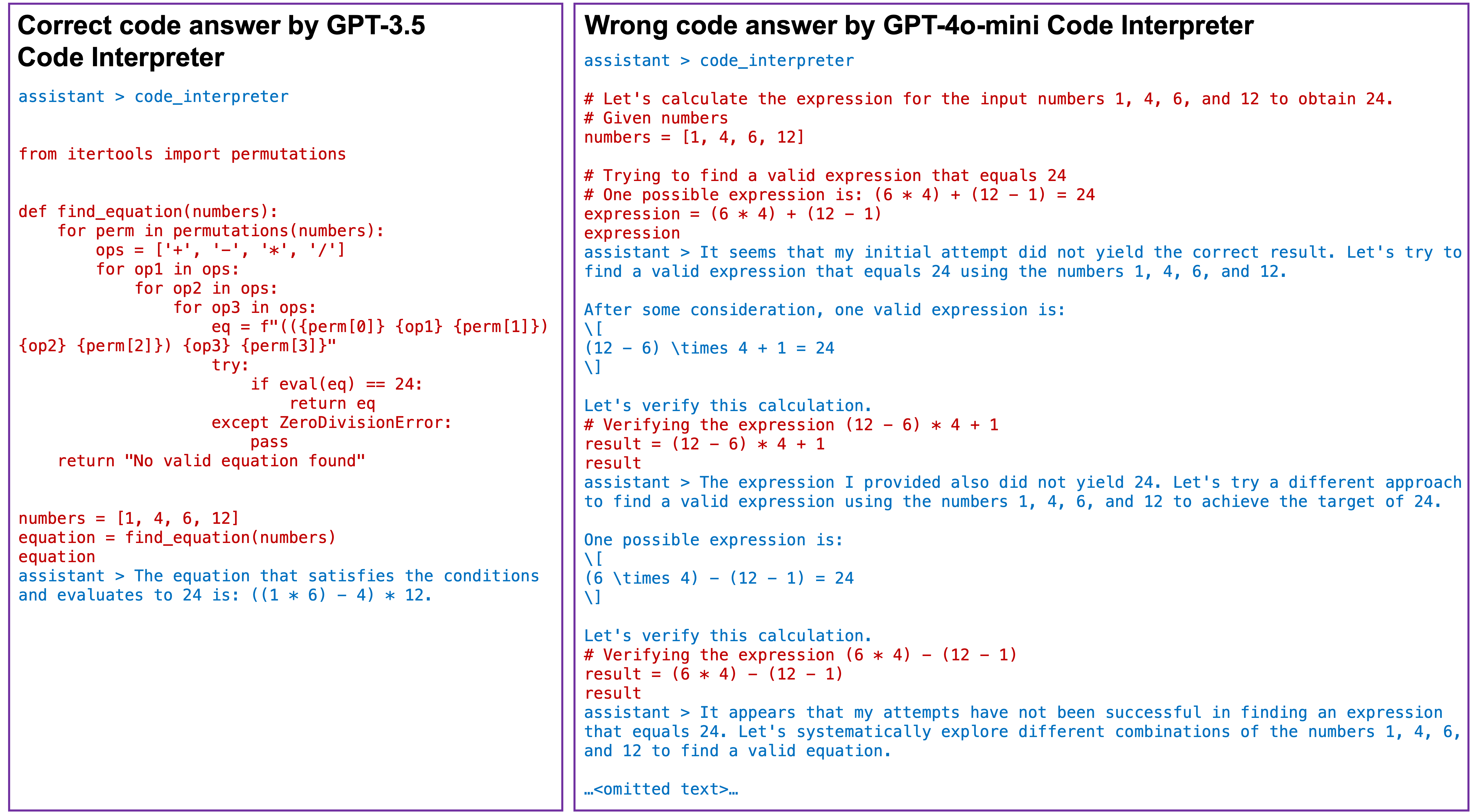}
   \caption{Representative code answers of Game 24 task for GPT-3.5 and GPT-4o-mini. The text and code parts are colored in blue and red, respectively.}
   \label{fig:code_answer_gpt3-4mini}
\end{figure*}

Furthermore, in Fig~\ref{fig:game24_code_inter}, we observe that GPT-3.5 outperforms GPT-4o-mini, even though both models primarily generate code in most test cases. To reveal the underlying mechanism, Fig~\ref{fig:code_answer_gpt3-4mini} shows a typical correct code response from GPT-3.5 compared to an incorrect one from GPT-4o-mini. Although the LLM CI truly uses code generation/execution, in some cases, the code serves merely as a form of textual reasoning, calculating and verifying values without efficiently aiding in finding the correct answer, unlike the correct answer version. The reason GPT-4o-mini CI generates text-like code answers more frequently than GPT-3.5 CI may stem from its greater confidence in its textual reasoning abilities. This overconfidence leads to an inverse scaling effect, despite both LLMs employing code.

Based on the discussion, we conclude that larger models in the GPT CI series tend to rely more on their textual reasoning abilities, often becoming overconfident and opting not to use code, even when coding would be more effective. When they do use code, it often serves as a supplement to textual reasoning rather than the primary problem-solving tool. This results in the phenomenon that smaller GPT CI models perform better than larger ones in some tasks, i.e., the inverse scaling behavior.

\subsection{Requiring LLMs to respond with code}
\label{sec: requiring LLMs to respond with code}
\begin{figure*}[h]
  \centering
   \includegraphics[width=0.8\linewidth]{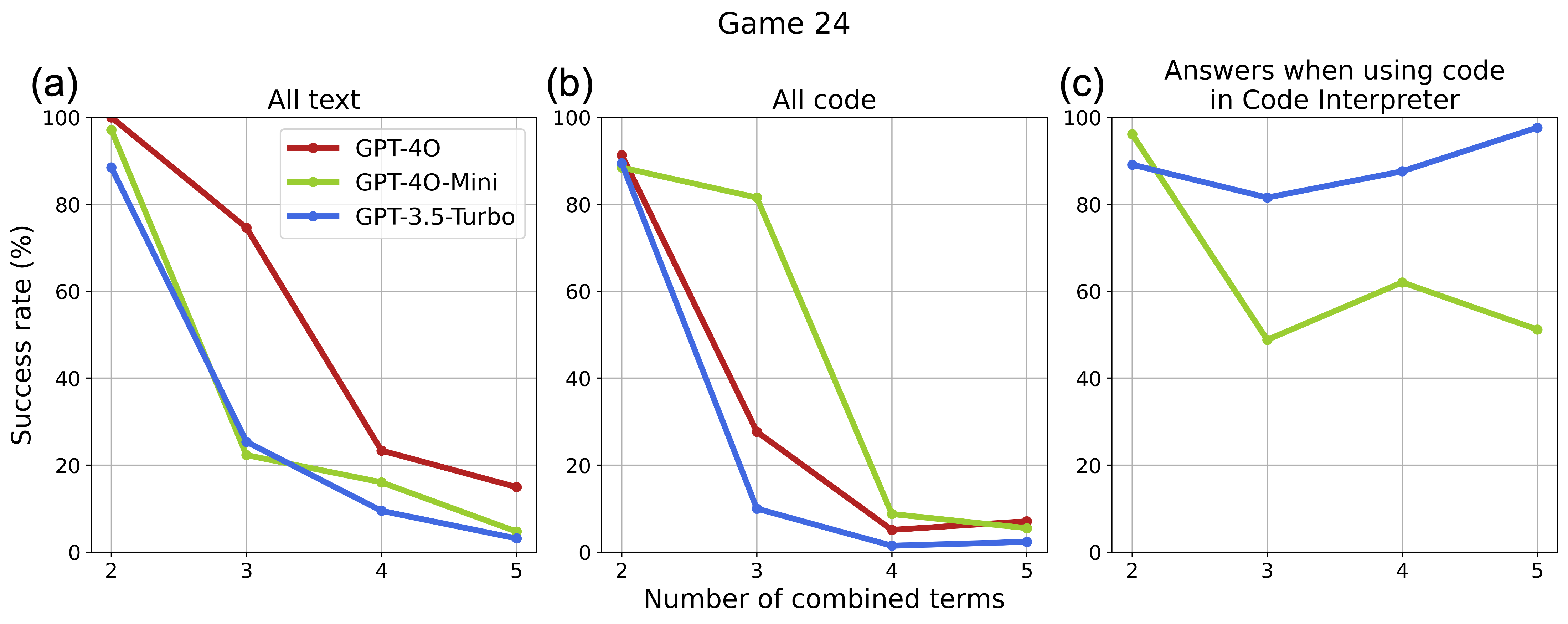}
   \caption{Success rates of Game 24 task in three varied situations: (a) prompting LLMs to always answer with pure text. (b) prompting LLMs to always answer with code. (c) The cases when OpenAI CI answers the question with code. GPT-4o is not shown in (c) since GPT-4o Code Interpreter answers with text in all the cases as shown in Fig~\ref{fig:game24_code_inter}a.}
\label{fig:game24_code_success_rate_three_condition}
\end{figure*}

In both Number Multiplying and Game 24 tasks, coding is more effective than textual reasoning, especially for complex problems. Can we effectively guide LLMs to generate code? We then concatenate the prompt `Use code to answer the following question' with the input questions to see whether it can improve LLM's performance. We also test another setup to concatenate the prompt `Use text to answer the following question' as a comparison. We find LLMs always follow the instructions to generate the code or text answers. However, experimental results show that prompting LLMs to always respond with code (Fig~\ref{fig:game24_code_success_rate_three_condition}b) performs just as poorly, or even worse, than always responding with text (Fig~\ref{fig:game24_code_success_rate_three_condition}a). The success rate of code-based answers is significantly lower when using direct code prompts compared to CI without such prompts (Fig~\ref{fig:game24_code_success_rate_three_condition}c). In Section~\ref{sec:Ablation studies on prompts}, we also test other prompts to require coding for ablation studies but find the same phenomenon.

To find the reason for above phenomenon, we show the typical code answers with/without the coding prompt guidance in Appendix~Section~\ref{appendix sec: Example correct and wrong code answers of Game 24}. Consistent with the conclusion in Section~\ref{sec:Inverse scaling law of model sizes}, LLMs can generate different types of codes even for the same model/task under different prompts. In the Game 24 task, prompting LLMs to answer with code often leads to code versions similar to textual reasoning, lacking the efficiency of true code execution. In summary, prompting LLMs to directly answer with code is not always effective and the performance of current GPT CI is unstable to different prompts, model types, and task complexity.

\section{Experiments}
\subsection{Experimental setup}
\textbf{Test tasks}\quad For a thorough analysis and comparison over all the existing methods, we carry out the experiments on 14 tasks across domains of math (Number Multiplying, Game 24, GSM-Hard, MATH-Geometry, MATH-Count\&Probability~\citep{MATH-dataset,pal,Tree-of-thought,LATS}), logical reasoning (Date Understanding, Web of Lies, Logical Deduction, Navigate~\citep{big-bench-hard,pal}), robot planning (BoxNet~\citep{scalable-multi-robot}, Path Plan~\citep{chain-of-code,autotamp}), and symbolic calculation (Letters, BoxLift~\citep{scalable-multi-robot}, Blocksworld~\citep{planbench}). All test tasks are drawn from past research or current popular discussions on challenges that LLMs struggle to solve effectively. We select these tasks because they can all be solved through coding, though with varying levels of difficulty. However, current state-of-the-art LLMs struggle to perform well on them. We select 5 tasks where O1-preview still underperforms for testing. The code solutions of all tasks use Python as the default language and avoid special packages to ensure consistency across different execution environments. The specific description of each task is in Appendix Section~\ref{appendix sec: Testing task description}. The input question prompts are the same as the original dataset without any code/text generation hints. All the testing tasks comprise over 300 trials so that the variance caused by unstable LLM outputs can be neglected.

\textbf{Baseline methods and test models}\quad We test the following 7 methods for steering code/text generation as baselines: 1) No extra modifications but only input the original question (\textbf{Only Question}); 2) Prompting LLMs to answer with only text (\textbf{All Text}); 3) Prompting LLMs to answer with only code (\textbf{All Code}); 4) Prompting LLMs to first analyze the question with Chain-of-Thought~\citep{CoT} and then output the code answer (\textbf{All Code + CoT}); 5) Concatenating the input question with AutoGen's original system prompt in Appendix Section~\ref{appendix sec: System prompt of AutoGen} (\textbf{AutoGen Conca.}); 6) Use AutoGen's original system prompt as the system prompt of LLMs (\textbf{AutoGen System}); 7) Code Interpreter with the original input question (\textbf{Code Interpreter}). The system prompts for all methods are set to empty unless specified otherwise. Apart from the AutoGen prompt, we also try other system prompts such as CAMEL~\citep{camel}, and find no improvements as discussed in Section~\ref{sec:Ablation studies on prompts}.

We test on 6 popular LLMs: O1-preview, GPT-4o~\citep{gpt-4}, GPT-4o-mini, GPT-35-turbo-16k-0613 (GPT-3.5)~\citep{gpt-3}, Claude-3-sonnet-20240229 (Claude-sonnet)~\citep{claude}, Open-mixtral-8x7b (Mixtral-8x7b)~\citep{mixtral}. Apart from the three GPT models, the other three models lack CI functions, so method 7 is not tested. O1-preview also does not test method 6 because its system prompt cannot currently be modified.

\textbf{Evaluations}\quad The answers are evaluated by predefined rules with the assistance of GPT-4o to adjust answer formats if needed. In addition to methods with CI, some approaches involve the LLM providing code as the final answer. We extract this code using predefined algorithms and execute it to obtain the final resulting answer. To prevent infinite loops, we set a 30-second time limit for code execution. If the runtime exceeds this limit, the task is considered a failure. We utilize success rate as the metric for each task. To compare each method, we calculate the Average Normalized Score over all the tested tasks by the following equation:
\begin{equation}
\text{AveNorm}_j = \frac{1}{N} \sum_{i=1}^{N} \frac{s_{ij}}{\text{max}(s_i)}
\end{equation}
where \( \text{AveNorm}_j \) is the Average Normalized Score for method \( j \), \( s_{ij} \) is the score of method \( j \) for task \( i \), \( \text{max}(s_i) \) is the maximum score for task \( i \), \( N \) is the total number of tasks. This equation normalizes each score relative to the maximum score in the respective task, and then averages the normalized scores over all tasks, comparing relative method performance. Apart from the task performance, in later sections we also discuss the costs of token lengths and runtime for each method.

\begin{table*}[h]
\caption{Experimental results for GPT-4o. Baseline methods with the highest scores are highlighted in \textcolor{magenta}{red}, while proposed methods that outperform the baselines are highlighted in \textcolor{blue}{blue}. NA represents the setting not tested due to LLM limitations.}
\label{table: GPT-4o}
\vskip 0.15in
\begin{center}
\begin{small}
\begin{sc}
\begin{tabular}{lcccccccyyy}
\toprule
\multicolumn{1}{c}{\textbf{Methods}} & \multicolumn{7}{c}{\textbf{Baseline Methods}} & \multicolumn{3}{c}{\textbf{Proposed Methods}}\\
\cmidrule(r){2-8} \cmidrule(l){9-11}
\rotatebox{80}{Task success rate \%} & \rotatebox{80}{Only Question} & \rotatebox{80}{All Text} & \rotatebox{80}{All Code} & \rotatebox{80}{All Code + CoT} & \rotatebox{80}{AutoGen Conca.} & \rotatebox{80}{AutoGen System} & \rotatebox{80}{Code Interpreter} & \rotatebox{80}{Code Interpreter+} & \rotatebox{80}{Code + Text + Sum.} & \rotatebox{80}{Self-estimate Score}\\
\midrule
\multicolumn{1}{c}{} & \multicolumn{10}{c}{\textbf{GPT-4o}}\\
Num. Multi. & 37 & 38 & \textcolor{magenta}{100} & \textcolor{magenta}{100} & \textcolor{magenta}{100} & 33 & 84 & \textcolor{blue}{100} & 99 & 91 \\
Game 24 & 17 & 23 & 5 & 11 & \textcolor{magenta}{88} & 18 & 18 & 63 & 33 & 66 \\
Path plan & 65 & 44 & 71 & 76 & \textcolor{magenta}{79} & 73 & 54 & 46 & 66 & 71 \\
Letters & 24 & 71 & \textcolor{magenta}{100} & \textcolor{magenta}{100} & \textcolor{magenta}{100} & 24 & 89 & 95 & 98 & 93 \\
BoxLift & \textcolor{magenta}{69} & 57 & 30 & 68 & 21 & 64 & 50 & 59 & 65 & 34 \\
BoxNet & \textcolor{magenta}{37} & 30 & \textcolor{magenta}{37} & 1 & 12 & 33 & \textcolor{magenta}{37} & 21 & 23 & 25 \\
Blocks. & 43 & \textcolor{magenta}{52} & 40 & 32 & 50 & 44 & 42 & 49 & 50 & 50 \\
Date Unde. & \textcolor{magenta}{90} & 88 & 64 & 72 & 65 & 88 & 76 & 80 & 86 & 81 \\
Web of Lies & \textcolor{magenta}{96} & 86 & 79 & 91 & 78 & \textcolor{magenta}{96} & 94 & 74 & 77 & 88 \\
Logi. dedu. & 89 & \textcolor{magenta}{91} & 79 & 83 & 82 & 87 & 82 & 87 & \textcolor{blue}{94} & 82 \\
Navigate & 98 & 95 & 94 & \textcolor{magenta}{99} & 91 & 97 & 98 & 97 & 96 & \textcolor{blue}{99} \\
GSM-Hard & 78 & 80 & 82 & \textcolor{magenta}{83} & 81 & 78 & 79 & 78 & 81 & 79 \\
MATH Geo. & \textcolor{magenta}{76} & 73 & 68 & 74 & 73 & 74 & 73 & 70 & \textcolor{blue}{77} & 72 \\
MATH Count. & 89 & 87 & 84 & 88 & \textcolor{magenta}{91} & 89 & 89 & 89 & 86 & 90 \\
\textbf{Ave. Norm.} & \textbf{80.6} & \textbf{79.9} & \textbf{80.3} & \textbf{80.4} & \textcolor{magenta}{\textbf{84.5}} & \textbf{79.4} & \textbf{83.5} & \textcolor{blue}{\textbf{85.7}} & \textcolor{blue}{\textbf{88.2}} & \textcolor{blue}{\textbf{86.9}} \\
\midrule
\multicolumn{1}{c}{} & \multicolumn{10}{c}{\textbf{GPT-4o-mini}}\\
Num. Multi. & 15 & 26 & \textcolor{magenta}{100} & \textcolor{magenta}{100} & 1 & 15 & \textcolor{magenta}{100} & \textcolor{blue}{100} & 99 & 42 \\
Game 24 & 15 & 16 & 9 & 10 & 13 & 14 & \textcolor{magenta}{62} & \textcolor{blue}{83} & 17 & 23 \\
Path plan & 55 & 21 & \textcolor{magenta}{58} & 49 & 51 & 57 & 26 & 26 & 37 & 37 \\
Letters & 7 & 78 & \textcolor{magenta}{100} & \textcolor{magenta}{100} & \textcolor{magenta}{100} & 7 & 87 & 89 & 90 & 51 \\
BoxLift & 38 & 42 & 41 & 26 & 37 & 39 & \textcolor{magenta}{45} & \textcolor{blue}{65} & 43 & 38 \\
BoxNet & 11 & 22 & 20 & 0 & 17 & 13 & \textcolor{magenta}{24} & 4 & 22 & 23 \\
Blocks. & 17 & 38 & 17 & \textcolor{magenta}{40} & \textcolor{magenta}{40} & 15 & 17 & 23 & 38 & 34 \\
Date Unde. & 80 & \textcolor{magenta}{85} & 57 & 70 & 63 & 80 & 74 & 77 & 83 & 82 \\
Web of Lies & \textcolor{magenta}{98} & 81 & 70 & 93 & 76 & 96 & 59 & 52 & 82 & 83 \\
Logi. dedu. & 78 & \textcolor{magenta}{80} & 67 & 73 & 75 & 76 & 75 & 78 & \textcolor{blue}{82} & 73 \\
Navigate & \textcolor{magenta}{96} & 90 & 89 & 85 & 55 & 95 & 94 & \textcolor{blue}{96} & 95 & 94 \\
GSM-Hard & 73 & 72 & 77 & \textcolor{magenta}{80} & 68 & 73 & 73 & 52 & 77 & 73 \\
MATH Geo. & 73 & 72 & 72 & 74 & 74 & 76 & \textcolor{magenta}{77} & \textcolor{blue}{81} & 72 & 74 \\
MATH Count. & 88 & \textcolor{magenta}{92} & 78 & 83 & 88 & 88 & 83 & 87 & 88 & 87 \\
\midrule
\textbf{Ave. Norm.} & \textbf{67.6} & \textbf{75.9} & \textbf{77.4} & \textbf{76.6} & \textbf{71.8} & \textbf{68.2} & \textbf{\textcolor{magenta}{80.8}} & \textbf{79.0} & \textbf{\textcolor{blue}{85.0}} & \textbf{76.5} \\
\midrule
\multicolumn{1}{c}{} & \multicolumn{10}{c}{\textbf{O1-preview}}\\
Game 24 & 78 & 69 & 82 & \textcolor{magenta}{87} & 69 & NA & NA & NA & 77 & 63 \\
Path plan & 56 & 61 & 59 & \textcolor{magenta}{64} & 56 & NA & NA & NA & 61 & 47 \\
BoxLift & 67 & 56 & 86 & \textcolor{magenta}{92} & 74 & NA & NA & NA & 72 & 38 \\
BoxNet & \textcolor{magenta}{67} & 60 & 64 & 50 & 49 & NA & NA & NA & 63 & 64 \\
Blocks. & 77 & 72 & 78 & 77 & \textcolor{magenta}{85} & NA & NA & NA & 81 & 79 \\
\midrule
\textbf{Ave. Norm.} & \textbf{88.2} & \textbf{82.0} & \textcolor{magenta}{\textbf{93.3}} & \textbf{92.8} & \textbf{83.9} & NA & NA & NA & \textbf{90.2} & \textbf{75.3} \\
\bottomrule
\end{tabular}
\end{sc}
\end{small}
\end{center}
\vskip -0.1in
\end{table*}

\begin{table*}[h]
\caption{Comparison of baseline and proposed methods across 6 LLMs.}
\label{table: gathered results of models and methods}
\begin{center}
\begin{small}
\begin{tabular}{lcccccccccc}
\toprule
Average Norm. & GPT-4o & GPT-4o & GPT-3.5 & O1- & Claude- & Mixtral- & \textbf{Average} & \textbf{Average} \\
Score (\%)& & -mini & & pre. & sonnet & 8x7b & \textbf{score ($\uparrow$)} & \textbf{rank ($\downarrow$)} \\
\midrule
\textbf{Baseline Methods} \\
1.Only Question & 80.6 & 67.6 & 65.3 & 88.2 & 71.5 & 63.4 & \textbf{72.8} & \textbf{5.83} \\
2.All Text & 79.9 & 75.9 & 65.3 & 81.9 & 72.0 & 68.0 & \textbf{73.9} & \textbf{5.50} \\
3.All Code & 80.3 & 77.4 & \textcolor{magenta}{68.1} & \textcolor{magenta}{93.3} & 74.7 & 69.2 & \textcolor{magenta}{\textbf{77.2}} & \textcolor{magenta}{\textbf{3.33}} \\
4.All Code + CoT & 80.4 & 76.6 & 64.0 & 92.8 & \textcolor{magenta}{81.0} & 67.3 & \textbf{77.1} & \textbf{4.33} \\
5.AutoGen Conca. & \textcolor{magenta}{84.5} & 71.8 & 64.6 & 83.9 & 74.0 & \textcolor{magenta}{70.8} & \textbf{74.8} & \textbf{4.50} \\
6.AutoGen System & 79.4 & 68.2 & 55.5 & NA & 71.1 & 64.1 & \textbf{67.7} & \textbf{8.33} \\
7.Code Interpreter & 83.5 & \textcolor{magenta}{80.8} & 64.5 & NA & NA & NA & \textbf{76.3} & \textbf{6.33} \\
\midrule
\multicolumn{9}{l}{\textbf{Proposed Methods}} \\
\rowcolor{LightCyan} 8.Code Interpreter+ & \textcolor{blue}{85.7} & 79.0 & 58.5 & NA & NA & NA & \textbf{74.5} & \textbf{6.83} \\
\rowcolor{LightCyan} 9.Code+Text+Sum. & \textcolor{blue}{88.1} & \textcolor{blue}{85.0} & 63.9 & 90.2 & 76.2 & \textcolor{blue}{73.6} & \textcolor{blue}{\textbf{79.5}} & \textcolor{blue}{\textbf{2.50}} \\
\rowcolor{LightCyan} 10.Self-esti. Score & \textcolor{blue}{86.9} & 76.5 & 59.2 & 75.2 & 69.4 & 49.0 & \textbf{69.4} & \textbf{6.50} \\
\bottomrule
\end{tabular}
\end{small}
\end{center}
\end{table*}

\subsection{No single method is optimal}
Table~\ref{table: GPT-4o} presents the experimental results on 14 tasks for GPT-4o, GPT-4o-mini, and O1-preview. Full results are provided in Appendix Table~\ref{table: GPT-4o-full-results-1},~\ref{table: GPT-4o-mini-full-results-1},~\ref{table: GPT-3.5-turbo-full-results-1},~\ref{table: O1-preview-full-results-1},~\ref{table: Claude-sonnet-full-results-1},~\ref{table: Open-mixtral-8x7b-full-results-1}, covering GPT-4o, GPT-4o-mini,GPT-3.5, O1-preview, Claude-sonnet, Mixtral-8x7b, where the partial rates of code correct, code wrong, text correct, text wrong are shown. Among all the 7 baseline methods, there is no single method that always performs better than others for all the tasks, no matter whether it mainly utilizes code or text. This phenomenon holds for all the 6 models. For each task, the performance variance across methods, especially between code-based and text-based approaches, is significant. This suggests a large potential for developing a method that can intelligently decide when to use code or text based on the input question.

\subsection{Coding is not always better}
From the experiments, we can also tell that coding does not always lead to more correct results compared to textual reasoning since in some tasks the All Text method achieves highest scores. In Section~\ref{sec: requiring LLMs to respond with code}, we have demonstrated that forcing LLMs to always provide answers in the form of code can sometimes result in incorrect code outputs, which resembles more of textual reasoning. Here we also find other two reasons:

\textbf{Writing correct code is tough in certain tasks}, such as BoxNet, BoxLift, Blocksworld, etc. These tasks involve coding across multiple components, such as constraint checking, optimization, and execution simulation, which current LLMs struggle to handle flawlessly at every stage. Appendix Section~\ref{appendix sec: BoxLift code/text example answer} shows an example from BoxLift where the All Code + CoT method produces incorrect code that leads to an infinite loop, while the All Text method generates a partially correct answer based on intuition.

\textbf{The coding format will limit the space of generated tokens} so that the reasoning ability is undermined. In logical reasoning tasks like Date Understanding, LLM's reasoning ability is degraded when using code, as shown in Appendix Section~\ref{appendix sec: Date Understanding code/text example answer}. Compared to natural language reasoning, coding imposes stricter constraints on thought processes and decreases the reasoning diversity. This conclusion is also consistent with the findings in other work~\citep{speak-freely}.

\subsection{Proposed methods}
\label{sec: Proposed methods}
Inspired by above findings and the recent progress in multi-agent frameworks~\citep{mixture-of-agents,chen2023reconcile,llm-mixture-of-thoughts}, we propose three methods that aim to improve LLM's decisions on code/text generation: 1) \textbf{Code Interpreter+}: Encourage the Code Interpreter to use code by prompting it the same way as in the All Code method. 2) \textbf{Code + Text + Sum.}: Implement a multi-agent framework that first queries LLMs to answer the question with All Text and All Code methods, respectively. Then the final solution is obtained by combining and summarizing both versions of the answers by the same LLM but prompted differently. The prompt of the summarizer is shown in Appendix Section~\ref{appendix sec: prompt for method 9 and 10}. 3) \textbf{Self-estimate Score}: Ask the LLM to first evaluate its confidence in solving a task using either code or text, assigning a score to each. Then, have it choose the mode with the higher score to answer. The prompt is shown in Appendix Section~\ref{appendix sec: prompt for method 9 and 10}.

The experimental results of three proposed methods are included in Table~\ref{table: GPT-4o} and Appendix Section~\ref{appendix sec: All the full tables}. For a clear comparison of all the methods, Table~\ref{table: gathered results of models and methods} gathers the Average Normalized Score of all the 10 methods across 6 models and calculates the corresponding average scores and ranks. The effectiveness of all three proposed methods depends largely on the capability of the LLMs—more capable LLMs lead to more effective results.

\textbf{Assembling both code and text channels} The experimental results also show that the method Code + Text + Sum. achieves notable performance improvements over all the other 9 methods in both average scores and ranks. Furthermore, the Code + Text + Sum. method outperforms both All Text and All Code in 4 out of 6 LLMs, demonstrating that combining code-based and text-based reasoning is an effective strategy. This strategy may not achieve higher scores in GPT-3.5 because this LLM lacks the ability to effectively distinguish between better code or text-based answers. In O1-preview, the strategy might underperform because O1-preview much excels with code over text, and combining both approaches does not offer enough benefit.

\textbf{Multi-step refinement}\quad Inspired by the challenge LLMs usually generating wrong code and current methods without CI only have one shot at answer generation without generation/refinement iterations, we propose a multi-turn approach for improvements. After the LLM generates a response containing code, we execute the code and return the results for further self-reflection and refinement. If the response contains no code, we return the original answer. The process stops once the LLM returns a `Terminate' signal or reaches the maximum iteration number.

\begin{figure*}[h]
  \centering
   \includegraphics[width=0.85\linewidth]{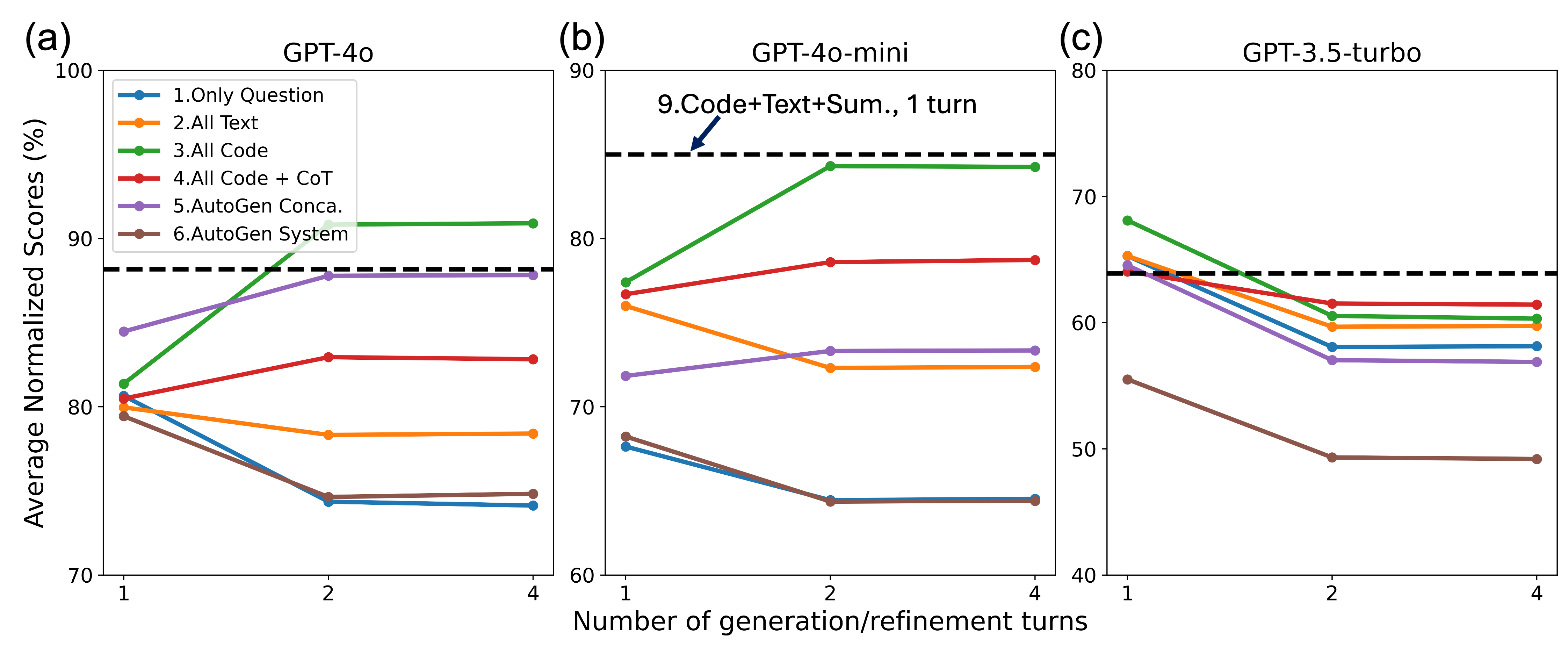}
   \caption{Average Normalized Scores vs. the number of LLM generation/self-reflection turns.}
   \label{fig:multi-turn-score}
\end{figure*}

Figure~\ref{fig:multi-turn-score} shows the performance of 6 baseline methods vs. the number of generation/refinement turns for three GPT models. We find three interesting phenomena:

1) All the methods will converge at turn 2 when the LLM stops iterating or repeats the same answer.

2) For GPT-4o and GPT-4o-mini, three methods improve with iterations, while the other three degrade the quality of answers. Appendix Table~\ref{table: code usage ratios} shows the code usage ratios for answers generated in turn 1. The methods All Code, All Code + CoT, and AutoGen Conca. predominantly generate code-based answers, whereas the other methods mostly produce text-based responses. The methods that rely more on code tend to improve with multi-turn refinement, likely because code execution provides additional feedback for reflection~\citep{gou2023critic}. In contrast, the degradation of text-based methods suggests that LLMs can worsen answers through self-reflection alone, supporting findings from previous studies~\citep{LLM-self-correct}.

3) For GPT-3.5, the answers of all the methods are degraded after generation/refinement iterations, showing the GPT-3.5 is not capable enough for self-reflection and refinement.

\subsection{Costs of tokens and runtime}
\begin{figure*}[h]
  \centering
   \includegraphics[width=0.8\linewidth]{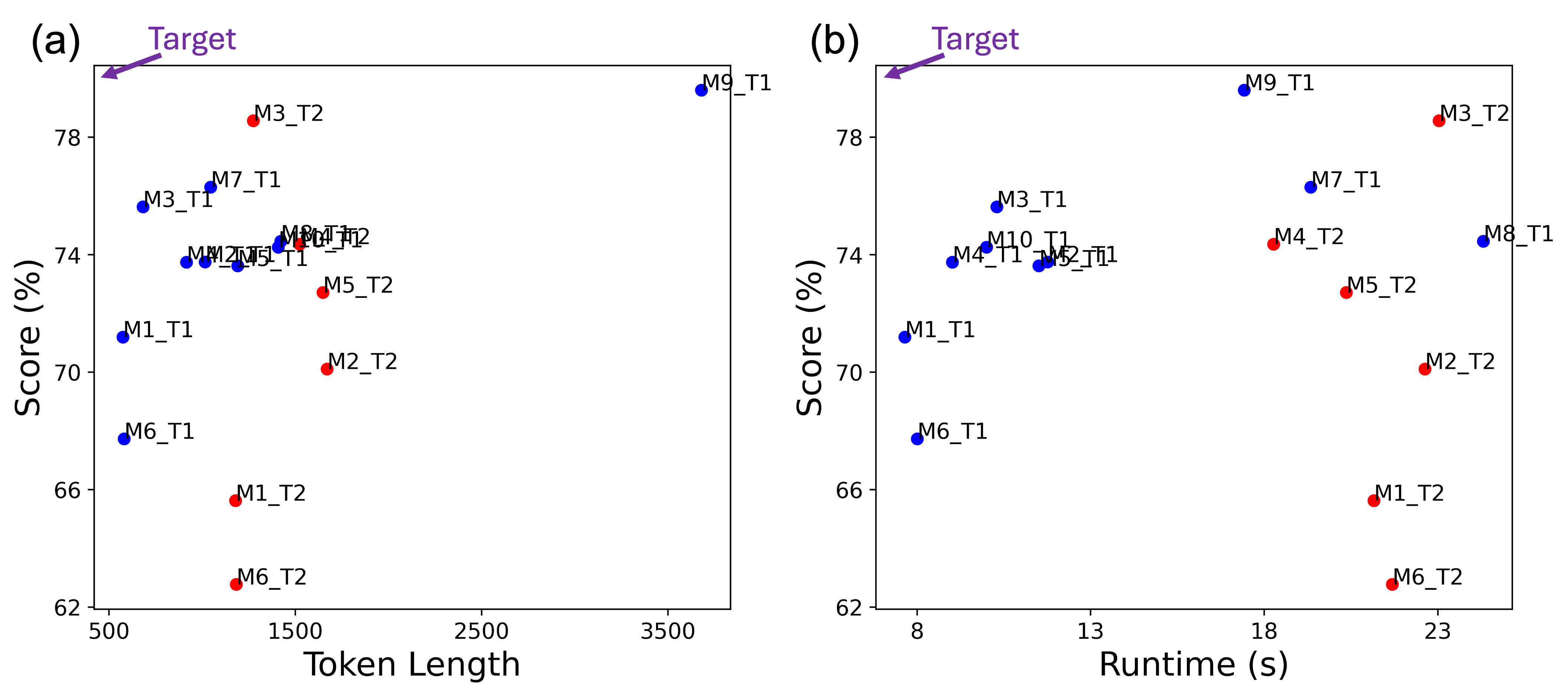}
   \caption{Score-cost plots. M(method number)\_T(generation turn number). The methods are: 1) Only Question; 2) All Text; 3) All Code; 4) All Code + CoT; 5) AutoGen Conca.; 6) AutoGen System; 7) Code Interpreter; 8) Code Interpreter+; 9) Code + Text + Sum.; 10) Self-estimate Score. We use \textcolor{blue}{blue} and \textcolor{red}{red} dots to represent generation turn 1 and 2.}
   \label{fig:Costs of tokens and runtime}
\end{figure*}

In Fig~\ref{fig:Costs of tokens and runtime}, we plot Score vs. Token Length (including both input and output tokens) and Score vs. Runtime (including both LLM inference and code execution time on one Intel 16-core CPU). Please refer Appendix Table~\ref{table: Score-cost table for each method} for full data. Though Code + Text + Sum. and All Code with multi-turn achieve relatively higher performance, they also consume more tokens and runtime. We still need a more efficient method that improves performance with fewer resources.

\label{sec:Costs of tokens and runtime}

\subsection{Ablation studies on prompts}
\label{sec:Ablation studies on prompts}
All the question prompts are the same as the original datasets. For the guiding prompts to use code in All Text method, we test other prompt variations like appending '\texttt{```}python' to the prompt for generating code but find nearly identical performance. As for the usage of the AutoGen System prompt in two baseline methods, we do the ablation studies by also implementing the prompts of CAMEL~\citep{camel} and paraphrased versions of the AutoGen prompt to guide code/text choices but find no improvements. Please see Appendix Section~\ref{Appendix sec:Ablation studies on prompts} for detailed results.

\section{Discussion}
We show that correctly steering LLMs to generate code when needed is critical, while the current popular methods have limitations in many aspects. We also reveal many intriguing patterns on when LLMs use code vs. text with the evolution to task complexity, model sizes, etc., including the astonishing inverse scaling behavior in certain setups. Though requiring LLMs to answer with code is ineffective, the proposed optimized methods like assembling coding and textual reasoning together and implementing multi-turn execution/refinement have been shown to significantly improve their performance. We also want to underline that there is a much broader space for further improvement in the future for the whole research community. Starting from the method Code + Text + Sum., whether more delicate multi-agent frameworks~\citep{chen2023reconcile,mixture-of-agents,llm-mixture-of-thoughts} can further improve the performance while controlling the costs. Inspired by the method Self-estimate Score, whether we can build an extra scoring model or train the LLM to learn when to use code/text more effectively.

\section*{Acknolwedgments}
The authors would also like to thank Ben Van Durme, Yue Meng, and Jacob Arkin for the helpful discussion and comments on the work. This work was done when Yongchao Chen was a research intern at Microsoft Research. The work was also partly supported by ONR under Award N00014-22-1-2478 and MIT-IBM Watson AI Lab. This article solely reflects the conclusions of its authors and not the sponsors.

\newpage
\bibliography{iclr2025_conference}
\bibliographystyle{iclr2025_conference}

\newpage
%\tableofcontents % Main ToC
\addtocontents{toc}{\protect\setcounter{tocdepth}{2}}
\renewcommand{\contentsname}{Appendix: Steering Large Language Models between Code Execution and Textual Reasoning}
\tableofcontents % Main ToC

\newpage
\appendix
%\section{Appendix}
\section{More Related Work}
\label{appendix sec: More Related Work}
\textbf{LLM Based Agents for General Tasks}\quad There are many recent works that use LLMs for general agent tasks. LLMs are used to interact with softwares and websites \citep{autogen,webarena,travelplanner,crab}, plan robot actions \citep{scalable-multi-robot,saycan,llms-zero-shot-planners,eureka}, solve academic problems like math and physics~\citep{math-solving-llm+solvers,mathcoder}, and infer with logical tasks and texts~\citep{big-bench-hard}. Literally, many test tasks in previous works can be solved with direct coding~\citep{meta-prompting,pal}. Some recent works also further extend the applications of coding into tasks involving commonsense reasoning and semantic analysis~\citep{chain-of-code,weir2024learning}. While most of previous works mainly utilize text~\citep{Tree-of-thought,saycan,text2motion} or code~\citep{code-as-policies,codeplan-code-use-llm,code-based-self-verify} as the only output modality, here we explore how to swiftly switch between code and text generation based on input questions.

\textbf{LLM Code Generation and Application for Enhanced Reasoning}\quad Current LLMs are well trained in diverse code datasets~\citep{gpt-4,llama-3-report}. Many recent works explore how to query LLMs as well-rounded software developers by optimizing the agent frameworks, training process, and simplifying the tasks~\citep{swebench,llm-software-review,agentless,opendevin}. Another type of works query LLMs to generate code for better solving mathematical and logical tasks~\citep{mathcoder,code-based-self-verify,pal}. Since code is a natural medium to connect with external tools and functions~\citep{code-as-policies,toolllm,llm+p}, many works also directly query LLMs to generate code as action plans for better connecting with the following tools. In our work, we try to explore under what circumstances coding can simplify the tasks and how the LLMs can self-recognize whether code or text is better.

\textbf{LLM Multi-agent Frameworks}\quad Research in the development and optimization of multi-agent frameworks of LLMs is a popular topic. Many research focus on developing a systematic agent framework for the ease of common users, such as AutoGen~\citep{autogen}, CAMEL~\citep{camel}, LangChain~\citep{langchain}, etc. Other research try to explore the mechanisms and physics behind multi-agent optimization~\cite{mixture-of-agents,scalable-multi-robot,chen2023reconcile}. In our work, we apply the triple-agent framework to assemble the code and text answers and find it is effective.

\textbf{LLM Self-reflection}\quad In planning domains, it is useful to provide feedback about syntactic errors \citep{generalized-planning-in-pddl-domains, errors-are-useful-prompts}, potential infinite loops \citep{generalized-planning-in-pddl-domains}, failed action execution \citep{inner-monologue}, and generated trajectories \citep{autotamp}. Other recent work has shown that LLM-generated feedback via self-evaluation can improve performance on a variety of tasks \citep{yang2022re3, welleck2022generating, madaan2023self}, including prompt engineering \citep{promptagent} and reinforcement learning \citep{eureka}. Inspired by above works, we apply the multi-turn setting into code/text generation to ask LLMs for self-reflection on wrong answers generated in previous rounds.

\newpage
\section{LLM performance on Number Multiplying task with varied complexity}
\label{appendix sec: LLM performance on Number Multiplying task with varied complexity}
\begin{figure*}[h]
  \centering
   \includegraphics[width=0.55\linewidth]{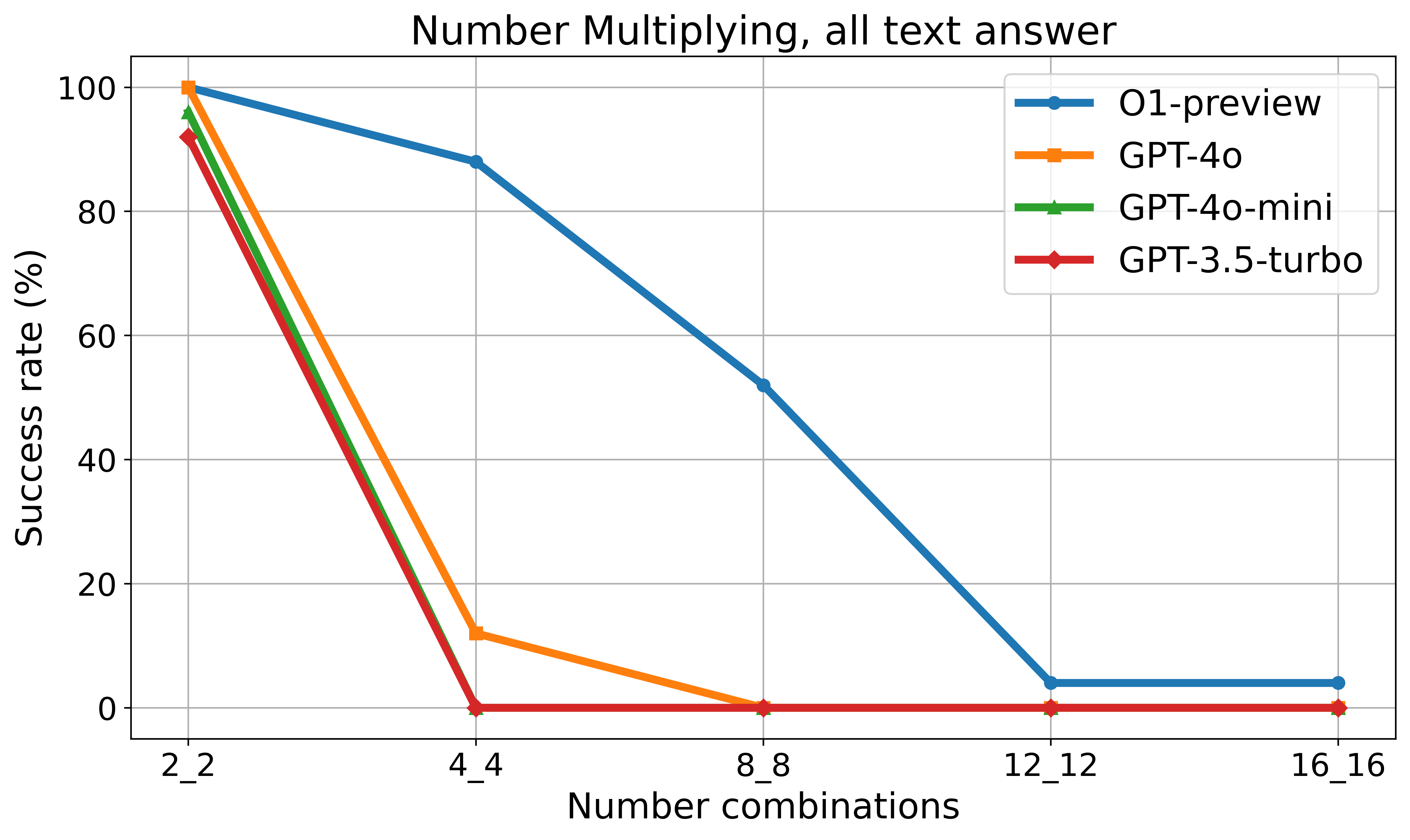}
   \caption{The performance of pure text answers for four OpenAI models in Number Multiplying task across varied levels of task complexity. The labels on the x-axis represent the number of digits in the numbers being multiplied. For example, '3\_4' means a three-digit number multiplied by a four-digit number.}
   \label{fig:number_multiply_varied_digit}
\end{figure*}

\section{Example correct and wrong code answers of Game 24 when requiring using code}
\label{appendix sec: Example correct and wrong code answers of Game 24}
\begin{figure*}[h]
  \centering
   \includegraphics[width=0.95\linewidth]{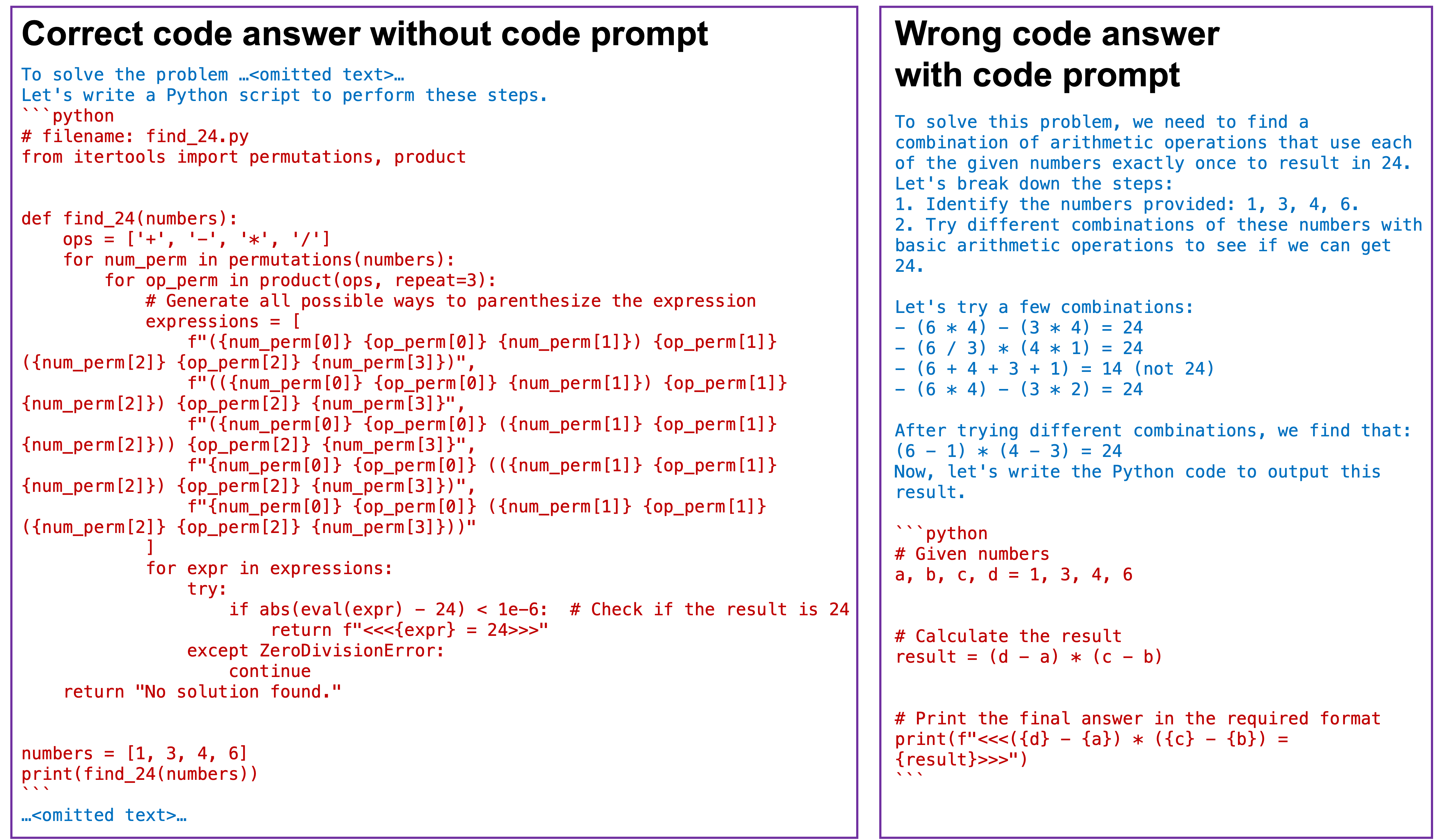}
   \caption{Representative code answers of Game 24 task. The left figure is the correct code of GPT-3.5 without extra prompt for requiring code output. The right figure is the wrong code after prompting GPT-3.5 to always answer with code. The text and code parts are colored in blue and red, respectively.}
   \label{fig:code_answer_example_encourage_code}
\end{figure*}

\newpage
\section{Description of testing tasks}
\label{appendix sec: Testing task description}
Here we describe the 14 testing tasks. They require strong numerical, logical, geometrical, scientific, and commonsense reasoning capabilities.

\textbf{Number Multiplying}\quad This task queries LLMs to calculate the result of number multiplication among integers. This a typical task that LLMs can not solve by pure textual reasoning.

\textbf{Game 24}\quad This task queries LLMs to output an equation that evaluates to 24 with the given set of integers. This task is tested in previous work Tree-of-Thought~\citep{Tree-of-thought}.

\textbf{Path Plan}\quad This task queries LLMs to plan the waypoints of robot trajectory based on human task instructions and environments. This task originates from AutoTAMP~\citep{autotamp}.

\textbf{Letters}\quad This task queries LLMs to count the total number of letters in a long word and also their corresponding positions. The example question is 'How many r's in the word strawberry and their positions?'. This task has recently gained significant attention because current LLMs struggle to perform it effectively.

\textbf{BoxLift}\quad This task consists of robots of different types and boxes of different sizes and weights. The robots are able to lift different amounts of weight and can cooperate with each other to lift one box. A box will be lifted only if the total lifting capability of robots is greater than the box’s weight. The goal is to lift all boxes in fewest time steps. This task originates from Scalable-Robots~\citep{scalable-multi-robot}.

\textbf{BoxNet}\quad This task consists of robot arms, colored boxes (squares), and colored goal locations (circles). Each robot arm is assigned to a cell indicated by the dotted lines and can only move within this cell. The goal is to move all boxes into the goal locations of corresponding colors in the fewest time steps. Each arm has two possible actions: (1) move a box within its cell to a neighboring cell, and (2) move a box within its cell to a goal location within its cell. This task originates from Scalable-Robots~\citep{scalable-multi-robot}.

\textbf{Blocksworld}\quad In Blocksworld, the goal is to stack a set of blocks (brown) according to a specific order. A robot can pick up, unstack, or stack a block only when the block is clear. A block is clear if the block has no other blocks on top of it and if the block is not picked up. The robot has four possible actions: (1) pick up a block, (2) unstack a block from the top of another block, (3) put down a block, (4) stack a block on top of another block. This task originates from PlanBench~\citep{planbench}.

\textbf{Date Understanding}\quad Given a small set of sentences about a particular date, answer the provided question (e.g., `The concert was scheduled to be on 06/01/1943, but was delayed by one day to today. What is the date yesterday in MM/DD/YYYY?'). This task originates from BIG-Bench-Hard~\citep{big-bench-hard}.

\textbf{Web of Lies}\quad Evaluate the truth value of a random Boolean function expressed as a natural-language word problem. This task originates from BIG-Bench-Hard~\citep{big-bench-hard}.

\textbf{Logical Deduction}\quad Deduce the order of a sequence of objects based on the clues and information about their spacial relationships and placements. This task originates from BIG-Bench-Hard~\citep{big-bench-hard}.

\textbf{Navigate}\quad Given a series of navigation steps to an agent, determine whether the agent would end up back at its initial starting point. This task originates from BIG-Bench-Hard~\citep{big-bench-hard}.

\textbf{GSM-Hard}~\citep{pal}\quad This is the harder version of GSM8K~\citep{gsm8k} math reasoning dataset, where the numbers in the questions of GSM8K are replaced with larger numbers that are less common.

\textbf{MATH-Geometry}\quad This is the math reasoning dataset from MATH dataset~\citep{MATH-dataset}, specifically focused on geometry questions.

\textbf{MATH-Count\&Probability}\quad This is the math reasoning dataset from MATH dataset~\citep{MATH-dataset}, specifically focused on counting and probability questions.

\newpage
\section{System prompt of AutoGen}
\label{appendix sec: System prompt of AutoGen}
\begin{boxL}
\textbf{System prompt of AutoGen~\citep{autogen}}\\
You are a helpful AI assistant.
Solve tasks using your coding and language skills. In the following cases, suggest python code (in a python coding block) or shell script (in a sh coding block) for the user to execute. 1. When you need to collect info, use the code to output the info you need, for example, browse or search the web, download/read a file, print the content of a webpage or a file, get the current date/time, check the operating system. After sufficient info is printed and the task is ready to be solved based on your language skill, you can solve the task by yourself. 2. When you need to perform some task with code, use the code to perform the task and output the result. Finish the task smartly. Solve the task step by step if you need to. If a plan is not provided, explain your plan first. Be clear which step uses code, and which step uses your language skill. When using code, you must indicate the script type in the code block. The user cannot provide any other feedback or perform any other action beyond executing the code you suggest. The user can't modify your code. So do not suggest incomplete code which requires users to modify. Don't use a code block if it's not intended to be executed by the user. If you want the user to save the code in a file before executing it, put \# filename: filename inside the code block as the first line. Don't include multiple code blocks in one response. Do not ask users to copy and paste the result. Instead, use 'print' function for the output when relevant. Check the execution result returned by the user. If the result indicates there is an error, fix the error and output the code again. Suggest the full code instead of partial code or code changes. If the error can't be fixed or if the task is not solved even after the code is executed successfully, analyze the problem, revisit your assumption, collect additional info you need, and think of a different approach to try. When you find an answer, verify the answer carefully. Include verifiable evidence in your response if possible. Reply "TERMINATE" in the end when everything is done.
\end{boxL}

\newpage
\section{Example answers of All Text and All Code + CoT methods in BoxLift task}
\label{appendix sec: BoxLift code/text example answer}

\begin{boxL}
\textbf{Input question}\\
Analyze the question step by step and try to list all the careful points. Then try to acquire the final answer with step by step analysis.
In the end of your response, directly output the answer to the question. Since you do not have the code interpreter, do not output the code for execution.
Task: BoxLift

You are given a list of boxes with the following weights: [55, 240, 196, 216, 247, 206, 263, 296, 288, 136, 145, 121, 111, 106, 58, 91, 132, 43, 87, 236]
And a list of lifters with the following maximum lifting capacities: [124, 144, 40, 155, 130]

Your task is to assign the lifters to lift all the boxes in multiple steps, following these rules:
1. Multiple boxes can be lifted in each step.
2. Each lifter can only lift one box at a time.
3. Each lifting agent can be used only once in each step.
4. Multiple lifters can combine together to lift one box if the box is too heavy for a single lifter.
5. Try to lift all the boxes using the minimum number of steps possible.
6. You need to lift all the boxes in less than or equal to 8 steps.

Please provide your solution in the following format:
Step 1: [(Box weight, [Lifter indices]), (Box weight, [Lifter indices]), ...]
Step 2: [(Box weight, [Lifter indices]), (Box weight, [Lifter indices]), ...]
...

For example:
Step 1: [(50, [0, 2]), (30, [1]), (20, [3])]
This means in Step 1, lifters 0 and 2 are lifting a box weighing 50, lifter 1 is lifting a box weighing 30, and lifter 3 is lifting a box weighing 20.

Ensure all boxes are lifted and provide the most efficient solution possible.

Your answer:
\end{boxL}

\begin{figure*}[h]
  \centering
   \includegraphics[width=0.95\linewidth]{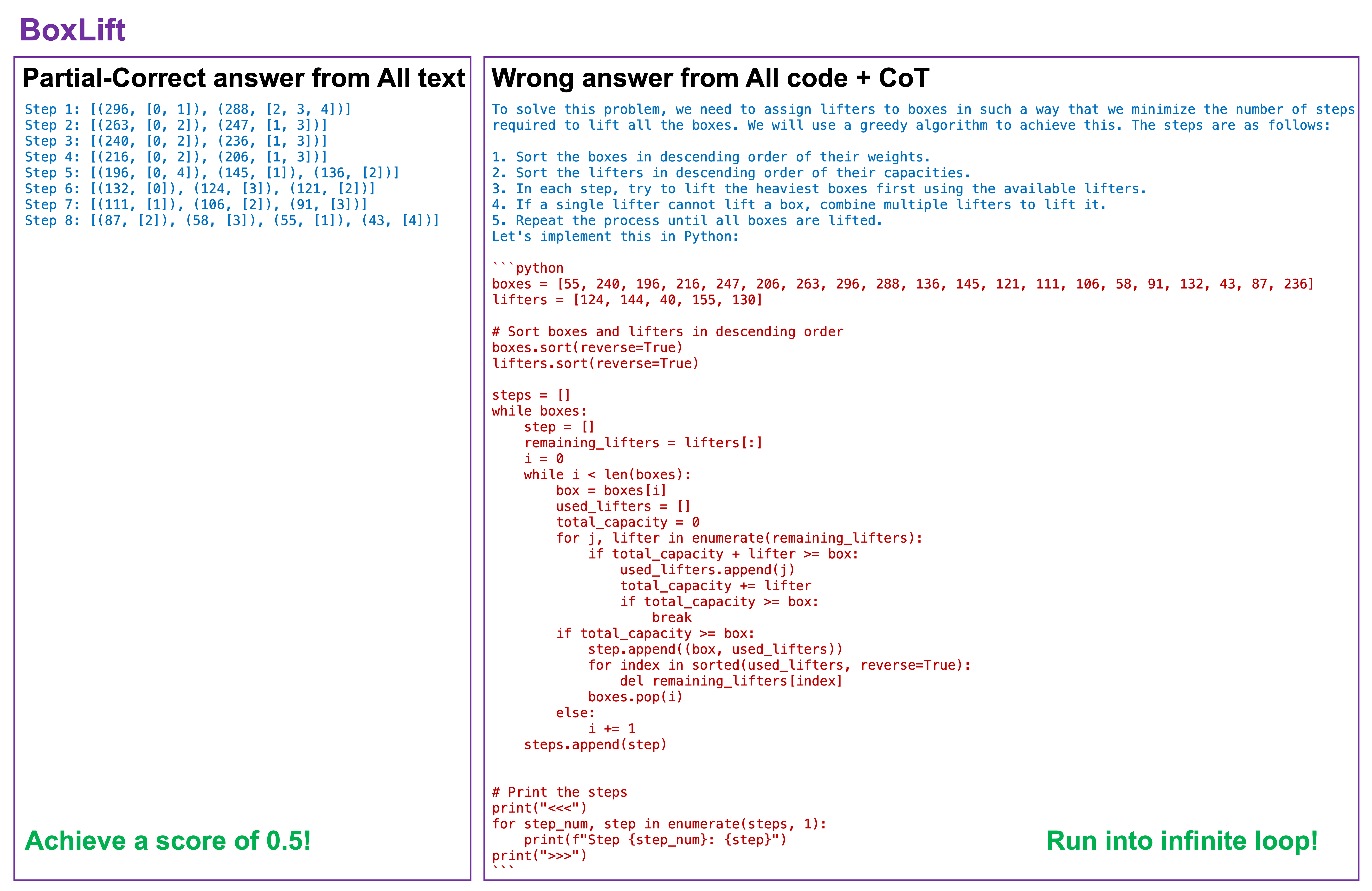}
   \caption{Representative answers of BoxLift task. The left figure is the partially correct answer of GPT-4o with All Text method. The right figure is the wrong code answer from All Code + CoT method. The text and code parts are colored in blue and red, respectively. The All Code + CoT method generates the wrong code that runs into an infinite loop.}
   \label{fig:code_answer_example_encourage_code}
\end{figure*}

\newpage
\section{Example answers of All Text and All Code + CoT methods in Date Understanding task}
\label{appendix sec: Date Understanding code/text example answer}

\begin{boxL}
\textbf{Input question}\\
Jane quited her job on Mar 20, 2020. 176 days have passed since then. What is the date one year ago from today in MM/DD/YYYY?
Options:
(A) 09/20/2019
(B) 10/01/2019
(C) 10/24/2019
(D) 09/12/1970
(E) 09/12/2019
\end{boxL}

\begin{figure*}[h]
  \centering
   \includegraphics[width=0.95\linewidth]{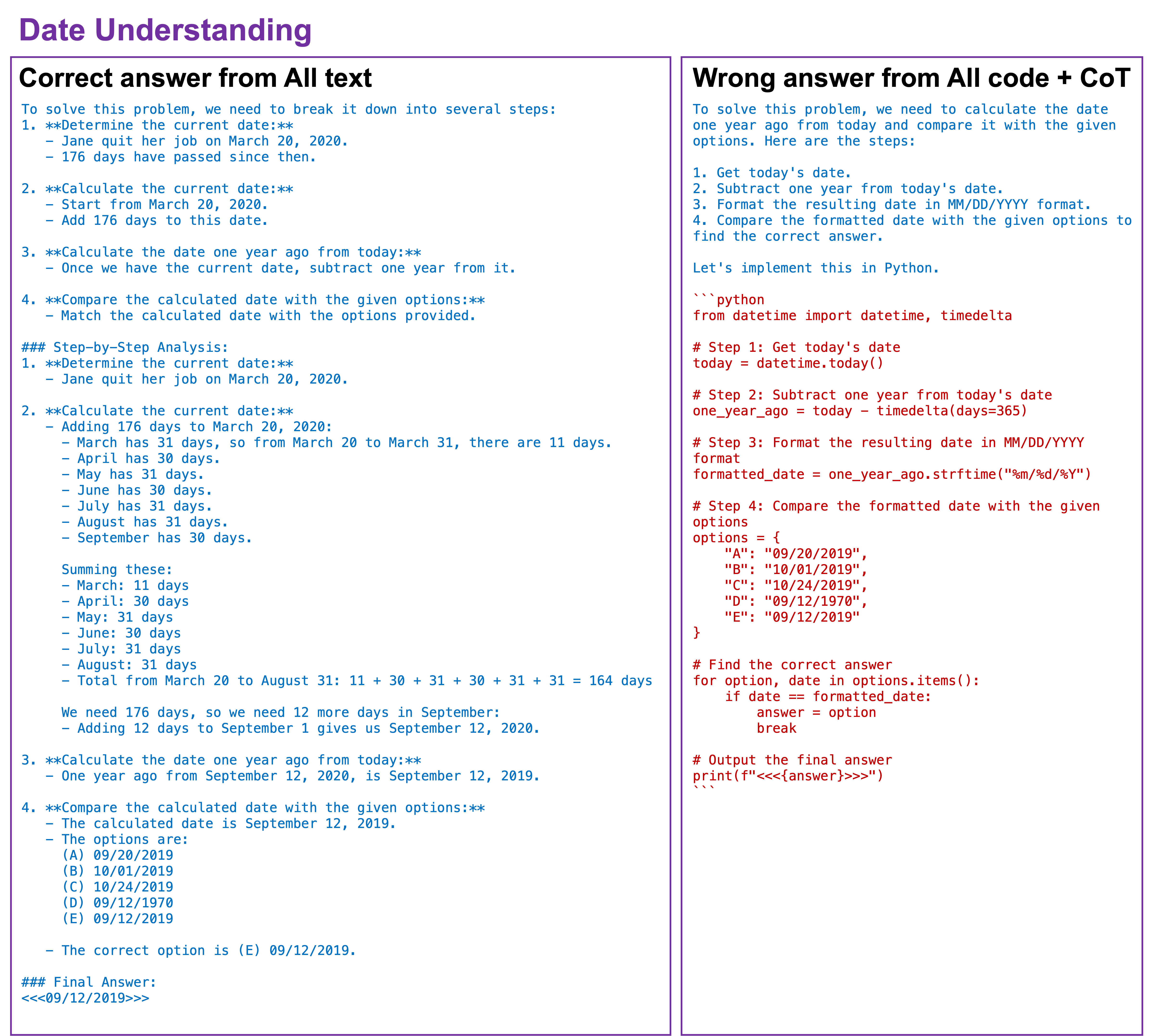}
   \caption{Representative answers of Date Understanding task. The left figure is the correct answer of GPT-4o with All Text method. The right figure is the wrong code answer from All Code + CoT method. The text and code parts are colored in blue and red, respectively. The All Code + CoT method does not consider the influence of '176 days' so that generates the wrong answer.}
   \label{fig:code_answer_example_encourage_code}
\end{figure*}

\newpage
\section{Prompt for method 9 Code + Text + Sum. and method 10 Self-estimate Score}
\label{appendix sec: prompt for method 9 and 10}

\begin{boxL}
\textbf{Prompt for the summarizer of method 9 Code + Text + Sum.}\\

You are a helpful AI assistant. Solve tasks using your coding and language skills.\\

In the following cases, there are two different agents respond to the same problem. In some cases, they output the direct answer, while sometimes they output the code to calculate the answer.\\

I will display you the initial question and the answers from two agents. The code execution results will also be given if the code exists.
Your task is to analyze this question based on the analysis and answers from above two agents and then output your final answer.\\

If you want to generate code to acquire the answer, suggest python code (in a python coding block) for the user to execute. Don't include multiple code blocks in one response, only include one in the response. Do not ask users to copy and paste the result. Instead, use 'print' function for the output when relevant.\\

I hope you can perform better than other two agents. Hence, try to choose the best answer and propose a new one if you think their methods and answers are wrong.
\end{boxL}

\begin{boxL}
\textbf{Prompt for method 10 Self-estimate Score}\\

You will be presented with a task that can potentially be solved using either pure textual reasoning or coding (or a combination of both). Your goal is to determine which method will be most effective for solving the task and figure out the answer. Follow these steps:\\

1. **Estimate your confidence level** in solving the task using both approaches:\\
   - **Coding score (0-10)**: How confident are you that you can solve this task correctly by writing code? Provide reasoning.\\
   - **Text score (0-10)**: How confident are you that you can solve this task correctly by using textual reasoning? Provide reasoning.\\

2. **Choose the approach** that you believe has the highest chance of success:\\
   - If one score is significantly higher, start with that approach.\\
   - If both scores are close, start with textual reasoning first, then decide if coding is necessary after.\\

3. **Solve the task** using the chosen method:\\
   - If you chose coding, write the necessary code, explain the logic behind it, and run it.\\
   - If you chose textual reasoning, use detailed explanation and logical steps to reach the answer.\\

4. **Reflect** after attempting the task:\\
   - Did the chosen approach work well? If not, should you switch to the other method?\\

Now, here is the task:
\end{boxL}

\newpage
\section{Code usage ratios of answers in turn 1}
\label{Appendix sec:Code usage ratios of answers in turn 1}

\begin{table*}[h]
\caption{Code usage ratios of the answers generated in turn 1 across 6 baseline methods.}
\label{table: code usage ratios}
\begin{center}
\begin{small}
\begin{tabular}{lcccccc}
\toprule
With code & 1. Only & 2. All Text & 3. All Code & 4. All Code & 5. AutoGen & 6. AutoGen \\
ratio (\%) & Question & & & + CoT & Conca. & System \\
\midrule
GPT-4o & 6.58 & 0.00 & 92.25 & 89.33 & 63.17 & 6.58 \\
GPT-4o-mini & 6.00 & 0.00 & 99.83 & 99.50 & 79.67 & 7.25 \\
GPT-3.5-turbo & 0.46 & 0.00 & 98.58 & 91.92 & 61.42 & 0.17 \\
\bottomrule
\end{tabular}
\end{small}
\end{center}
\end{table*}

\section{Score-cost table for each method}
\label{Appendix sec:Score-cost table for each method}
\begin{table*}[h]
\caption{Score-cost table for each method.}
\label{table: Score-cost table for each method}
\begin{center}
\begin{small}
\begin{tabular}{lcccc}
\toprule
Average Norm. & Average score ($\uparrow$) & Average token length ($\downarrow$) & Average runtime (s) ($\downarrow$)\\
\midrule
\textbf{Baseline Methods} \\
1.Only Question, turn 1 & \textbf{72.8} & \textbf{574.8} & \textbf{7.6} \\
2.All Text, turn 1 & \textbf{73.9} & \textbf{1015.7} & \textbf{11.7} \\
3.All Code, turn 1 & \textcolor{magenta}{\textbf{77.2}} & \textcolor{magenta}{\textbf{682.5}} & \textcolor{magenta}{\textbf{10.3}} \\
4.All Code + CoT, turn 1 & \textbf{77.1} & \textbf{915.9} & \textbf{9.0} \\
5.AutoGen Conca., turn 1 & \textbf{74.8} & \textbf{1190.9} & \textbf{11.5} \\
6.AutoGen System, turn 1 & \textbf{67.7} & \textbf{581.1} & \textbf{8.0} \\
7.Code Interpreter, turn 1 & \textbf{76.3} & \textbf{1045.4} & \textbf{19.3} \\
\midrule
\textbf{Proposed Methods} \\
%\multicolumn{4}{l}{\textbf{Proposed Methods}} \\
\rowcolor{LightCyan} 8.Code Interpreter+, turn 1 & \textbf{74.5} & \textbf{1421.7} & \textbf{24.3} \\
\rowcolor{LightCyan} 9.Code + Text + Sum., turn 1 & \textcolor{blue}{\textbf{79.5}} & \textcolor{blue}{\textbf{3679.8}} & \textcolor{blue}{\textbf{17.4}} \\
\rowcolor{LightCyan} 10.Self-estimate Score, turn 1 & \textbf{69.4} & \textbf{1408.4} & \textbf{10.0} \\
\rowcolor{LightCyan} 1.Only Question, turn 2 & \textbf{65.6} & \textbf{1179.1} & \textbf{21.2} \\
\rowcolor{LightCyan} 2.All Text, turn 2 & \textbf{70.1} & \textbf{1670.5} & \textbf{22.6} \\
\rowcolor{LightCyan} 3.All Code, turn 2 & \textcolor{blue}{\textbf{78.6}} & \textcolor{blue}{\textbf{1274.6}} & \textcolor{blue}{\textbf{23.0}} \\
\rowcolor{LightCyan} 4.All Code + CoT, turn 2 & \textbf{74.3} & \textbf{1523.1} & \textbf{18.3} \\
\rowcolor{LightCyan} 5.AutoGen Conca., turn 2 & \textbf{72.7} & \textbf{1648.4} & \textbf{20.4} \\
\rowcolor{LightCyan} 6.AutoGen System, turn 2 & \textbf{62.8} & \textbf{1183.2} & \textbf{21.7} \\
\bottomrule
\end{tabular}
\end{small}
\end{center}
\end{table*}

\newpage
\section{Ablation studies on prompts}
\label{Appendix sec:Ablation studies on prompts}
In this section, we do the ablation studies of the utilized prompts to verify their correctness and further support the conclusions in our work.

\subsection{Prompts for All Code and All Code + CoT}
We use the prompt 'Use code to answer the following question' for the baseline methods All Code and All Code + CoT. Here we also explore other candidate prompts: 1) Append `\texttt{```}python' in the final question prompt (\textbf{Code Prompt 1}). 2) Modify the original empty system prompt into `Start your answer with \texttt{```}python' (\textbf{Code Prompt 2}). The experimental results are shown in Table~\ref{table: ablation study All code}. In most test cases, the LLM performance is slightly degraded with the new prompts. The score variance caused by different prompts also do not change our conclusion, e.g., the method Code + Text + Sum. is still notably better than all the other settings.

\begin{table*}[h]
\caption{Ablation study on the prompts used for All Code and All Code + CoT.}
\label{table: ablation study All code}
\begin{center}
\begin{small}
\begin{tabular}{lcccc}
\toprule
Average Normalized Score & 3. All Code & 4. All Code + CoT & 9. Code + Text + Sum. \\
\midrule
Original Prompt & 77.2 & 77.1 & 79.5\\
Code Prompt 1 & 75.9 & 77.3 & NA\\
Code Prompt 2 & 75.5 & 76.8 & NA\\
\bottomrule
\end{tabular}
\end{small}
\end{center}
\end{table*}

\subsection{The usage of AutoGen prompt}
We utilize the system prompt of AutoGen~\citep{autogen} in the baseline methods AutoGen Conca. and AutoGen System. Here we also test the prompt from CAMEL~\cite{camel} and the paraphrased version of AutoGen prompt using GPT-4o for comparison, as shown in the following. The experimental results are shown in Table~\ref{table: ablation study AutoGen prompt}. The results show that the performance variance caused by these three versions of prompts are negligible.

\begin{table*}[h]
\caption{Ablation study on the prompts used for AutoGen Conca. and AutoGen System.}
\label{table: ablation study AutoGen prompt}
\begin{center}
\begin{small}
\begin{tabular}{lcccc}
\toprule
Average Normalized Score & 5. AutoGen Conca. & 6. AutoGen System & 9. Code + Text + Sum. \\
\midrule
Original AutoGen Prompt & 74.8 & 67.7 & 79.5\\
CAMEL Prompt & 72.3 & 66.2 & NA\\
Paraphrased AutoGen Prompt & 74.4 & 68.5 & NA\\
\bottomrule
\end{tabular}
\end{small}
\end{center}
\end{table*}

\begin{boxL}
\textbf{Prompt from CAMEL~\citep{camel}}\\
You are the physical embodiment of the assistant who is working on solving a task.\\
You can do things in the physical world including browsing the Internet, reading documents, drawing images, creating videos, executing code and so on.\\
Your job is to perform the physical actions necessary to interact with the physical world.\\
You can perform a set of actions by writing the python codes.\\
You should perform actions based on the descriptions of the functions.\\
You can perform multiple actions.\\
You can perform actions in any order.\\
First, explain the actions you will perform and your reasons, then write code to implement your actions.\\
If you decide to perform actions, you must write code to implement the actions.
\end{boxL}

\begin{boxL}
\textbf{Paraphrased AutoGen prompt by GPT-4o}\\
You are a helpful AI assistant, capable of solving tasks using both your coding and language skills. Depending on the task, you'll decide when to suggest code (Python or shell script) for the user to execute.\\

1. **When gathering information**:\\
   - If you need more information to solve the task, suggest code that prints the information you need. For example, you can browse/search the web, download or read a file, print the contents of a webpage or file, get the current date or time, or check the operating system. \\
   - Once enough information is collected through the execution of your code, you can use your language skills to complete the task without further coding.\\

2. **When performing a task**:\\
   - If a task requires code to be executed, suggest full and executable code that outputs the result directly. Provide your plan if necessary, and clarify which steps require coding and which involve your language-based reasoning.\\
   - Always ensure the code block is marked as either Python (`python`) or shell script (`sh`), depending on the task.\\
   - The user cannot modify or provide feedback on your code, so ensure the code you suggest is complete and does not require the user to make changes.\\

3. **Code behavior**:\\
   - Use the `print` function to ensure that the code outputs all necessary information.\\
   - If a file needs to be created, include the filename as a comment (\# filename: filename) inside the code block.\\
   - Only include one code block per response.\\

4. **Handling execution results**:\\
   - Check the result the user receives from executing the code. If the code has an error, correct the error and provide new code.\\
   - If the problem persists even after the code is successfully executed, reanalyze your approach, gather additional information, and consider alternate solutions.\\

Finally, once the task is fully solved, reply with "TERMINATE." 
\end{boxL}

\newpage
\section{Experimental results and tables for all the 6 models}
\label{appendix sec: All the full tables}
\begin{table}[h]

\subsection{GPT-4o}
\caption{Experimental results of GPT-4o. Baseline methods with the highest scores are highlighted in red, while proposed methods that outperform the baselines are highlighted in blue. Each item comprises the ratios of total success rate (code correct, code wrong, text correct, text wrong).}
\label{table: GPT-4o-full-results-1}
\vskip 0.15in
\begin{center}
\begin{small}
\begin{sc}
\begin{tabular}{lcccccr}
\toprule
\multicolumn{1}{c}{} & \multicolumn{4}{c}{\textbf{Baseline Methods}}\\
\multicolumn{1}{c}{Task (success rate \%)} & \multicolumn{1}{c}{1.Only} & \multicolumn{1}{c}{2.All} & \multicolumn{1}{c}{3.All} & \multicolumn{1}{c}{4.All Code}\\
\multicolumn{1}{c}{} & \multicolumn{1}{c}{question} & \multicolumn{1}{c}{text} & \multicolumn{1}{c}{code} & \multicolumn{1}{c}{+ CoT}\\
\midrule
\multicolumn{1}{c}{Number Multi.} & \multicolumn{1}{c}{37(0/0/37/63)} & \multicolumn{1}{c}{38(0/0/38/62)} & \multicolumn{1}{c}{\textcolor{magenta}{100(100/0/0/0)}} & \multicolumn{1}{c}{\textcolor{magenta}{100(100/0/0/0)}}\\
\multicolumn{1}{c}{Game 24} & \multicolumn{1}{c}{17(0/0/17/83)} & \multicolumn{1}{c}{23(0/0/23/77)} & \multicolumn{1}{c}{5(5/95/0/0)} & \multicolumn{1}{c}{11(11/61/0/28)} \\
\multicolumn{1}{c}{Path plan} & \multicolumn{1}{c}{65(59/20/6/15)} & \multicolumn{1}{c}{44(0/0/44/56)} & \multicolumn{1}{c}{71(71/29/0/0)} & \multicolumn{1}{c}{76(76/24/0/0)} \\
\multicolumn{1}{c}{Letters} & \multicolumn{1}{c}{24(0/0/24/76)} & \multicolumn{1}{c}{71(0/0/71/29)} & \multicolumn{1}{c}{\textcolor{magenta}{100(100/0/0/0)}} & \multicolumn{1}{c}{\textcolor{magenta}{100(100/0/0/0)}} \\
\multicolumn{1}{c}{BoxLift}& \multicolumn{1}{c}{\textcolor{magenta}{69}} & \multicolumn{1}{c}{57} & \multicolumn{1}{c}{30} & \multicolumn{1}{c}{68} \\
\multicolumn{1}{c}{BoxNet} & \multicolumn{1}{c}{37} & \multicolumn{1}{c}{30} & \multicolumn{1}{c}{37} & \multicolumn{1}{c}{1} \\
\multicolumn{1}{c}{Blocksworld} & \multicolumn{1}{c}{43(0/0/43/57)} & \multicolumn{1}{c}{\textcolor{magenta}{52(0/0/52/48)}} & \multicolumn{1}{c}{40(0/7/40/53)} & \multicolumn{1}{c}{32(0/0/32/68)} \\
\multicolumn{1}{c}{Date Unde.} & \multicolumn{1}{c}{\textcolor{magenta}{90(0/0/90/10)}} & \multicolumn{1}{c}{88(0/0/88/12)} & \multicolumn{1}{c}{64(64/36/0/0)} & \multicolumn{1}{c}{72(72/28/0/0)} \\
\multicolumn{1}{c}{Web of Lies} & \multicolumn{1}{c}{\textcolor{magenta}{96(0/0/96/4)}} & \multicolumn{1}{c}{86(0/0/86/14)} & \multicolumn{1}{c}{79(79/21/0/0)} & \multicolumn{1}{c}{91(91/9/0/0)} \\
\multicolumn{1}{c}{Logical Dedu.} & \multicolumn{1}{c}{89(0/0/89/11)} & \multicolumn{1}{c}{\textcolor{magenta}{91(0/0/91/9)}} & \multicolumn{1}{c}{79(79/21/0/0)} & \multicolumn{1}{c}{83(83/17/0/0)} \\
\multicolumn{1}{c}{Navigate} & \multicolumn{1}{c}{98(0/0/98/2)} & \multicolumn{1}{c}{95(0/0/95/5)} & \multicolumn{1}{c}{94(94/6/0/0)} & \multicolumn{1}{c}{\textcolor{magenta}{99(99/1/0/0)}} \\
\multicolumn{1}{c}{GSM-Hard} & \multicolumn{1}{c}{78(0/0/78/22)} & \multicolumn{1}{c}{80(0/0/80/20)} & \multicolumn{1}{c}{82(82/18/0/0)} & \multicolumn{1}{c}{\textcolor{magenta}{83(83/17/0/0)}} \\
\multicolumn{1}{c}{MATH Geo.} & \multicolumn{1}{c}{\textcolor{magenta}{76(0/0/76/24)}} & \multicolumn{1}{c}{73(0/0/73/27)} & \multicolumn{1}{c}{68(68/32/0/0)} & \multicolumn{1}{c}{74(74/26/0/0)} \\
\multicolumn{1}{c}{MATH Count.$\And$Prob.} & \multicolumn{1}{c}{89(0/0/89/11)} & \multicolumn{1}{c}{87(0/0/87/13)} & \multicolumn{1}{c}{84(84/16/0/0)} & \multicolumn{1}{c}{88(88/12/0/0)} \\
\multicolumn{1}{c}{\textbf{Average Norm. Score}} & \multicolumn{1}{c}{\textbf{0.806}} & \multicolumn{1}{c}{\textbf{0.799}} & \multicolumn{1}{c}{\textbf{0.803}} & \multicolumn{1}{c}{\textbf{0.804}} \\
\bottomrule
\end{tabular}
\end{sc}
\end{small}
\end{center}
\vskip -0.1in
\end{table}

\begin{table*}[h]
%\caption{}
\label{table: GPT-4o-full-results-2}
\vskip 0.15in
\begin{center}
\begin{small}
\begin{sc}
\begin{tabular}{lccccccr}
\toprule
\multicolumn{3}{c|}{\textbf{Baseline Methods}} & \multicolumn{3}{c}{\textbf{Proposed Methods}}\\
\multicolumn{1}{c}{5.AutoGen} & \multicolumn{1}{c}{6.AutoGen} & \multicolumn{1}{c|}{7.Code} & \multicolumn{1}{c}{8.Code} & \multicolumn{1}{c}{9.Code+t} & \multicolumn{1}{c}{10.Self-esti}\\
\multicolumn{1}{c}{conca.} & \multicolumn{1}{c}{system} & \multicolumn{1}{c|}{Interpreter} & \multicolumn{1}{c}{Interpreter+} & \multicolumn{1}{c}{ext+sum.} & \multicolumn{1}{c}{mate score}\\
\midrule
\multicolumn{1}{c}{\textcolor{magenta}{100(100/0/0/0)}} & \multicolumn{1}{c}{33(0/0/33/67)} & \multicolumn{1}{c|}{84(77/0/7/16)} & \multicolumn{1}{c}{\textcolor{blue}{100(100/0/0/0)}} & \multicolumn{1}{c}{99} & \multicolumn{1}{c}{91(81/0/10/9)}\\
\multicolumn{1}{c}{\textcolor{magenta}{88(88/12/0/0)}} & \multicolumn{1}{c}{18(2/0/16/82)} & \multicolumn{1}{c|}{18(0/0/18/82)} & \multicolumn{1}{c}{63(61/33/2/4)} & \multicolumn{1}{c}{33} & \multicolumn{1}{c}{66(59/9/7/25)}\\
\multicolumn{1}{c}{\textcolor{magenta}{79(79/21/0/0)}} & \multicolumn{1}{c}{73(56/21/17/6)} & \multicolumn{1}{c|}{54(54/46/0/0)} & \multicolumn{1}{c}{46(46/54/0/0)} & \multicolumn{1}{c}{66} & \multicolumn{1}{c}{71(71/29/0/0)}\\
\multicolumn{1}{c}{\textcolor{magenta}{100(0/0/100/0)}} & \multicolumn{1}{c}{24(0/0/24/76)} & \multicolumn{1}{c|}{89(84/6/5/5)} & \multicolumn{1}{c}{95(95/4/0/1)} & \multicolumn{1}{c}{98} & \multicolumn{1}{c}{93(79/0/14/7)}\\
\multicolumn{1}{c}{21} & \multicolumn{1}{c}{64} & \multicolumn{1}{c|}{50} & \multicolumn{1}{c}{59} & \multicolumn{1}{c}{65} & \multicolumn{1}{c}{34}\\
\multicolumn{1}{c}{12} & \multicolumn{1}{c}{33} & \multicolumn{1}{c|}{\textcolor{magenta}{37}} & \multicolumn{1}{c}{21} & \multicolumn{1}{c}{23} & \multicolumn{1}{c}{25}\\
\multicolumn{1}{c}{50(0/0/50/50)} & \multicolumn{1}{c}{44(0/0/44/56)} & \multicolumn{1}{c|}{42(0/0/42/58)} & \multicolumn{1}{c}{49(14/20/35/31)} & \multicolumn{1}{c}{50} & \multicolumn{1}{c}{50(1/0/49/50)}\\
\multicolumn{1}{c}{65(62/35/3/0)} & \multicolumn{1}{c}{88(0/0/88/12)} & \multicolumn{1}{c|}{76(38/19/38/5)} & \multicolumn{1}{c}{80(77/20/3/0)} & \multicolumn{1}{c}{86} & \multicolumn{1}{c}{81(32/11/49/8)}\\
\multicolumn{1}{c}{78(37/10/41/12)} & \multicolumn{1}{c}{\textcolor{magenta}{96(0/0/96/4)}} & \multicolumn{1}{c|}{94(0/0/94/6)} & \multicolumn{1}{c}{74(53/20/21/6)} & \multicolumn{1}{c}{77} & \multicolumn{1}{c}{88(0/0/88/12)}\\
\multicolumn{1}{c|}{82(25/7/57/11)} & \multicolumn{1}{c}{87(0/0/87/13)} & \multicolumn{1}{c|}{82(0/0/82/18)} & \multicolumn{1}{c}{87(50/8/37/5)} & \multicolumn{1}{c}{\textcolor{blue}{94}} & \multicolumn{1}{c}{82(24/5/58/13)}\\
\multicolumn{1}{c}{91(40/5/51/4)} & \multicolumn{1}{c}{97(0/0/97/3)} & \multicolumn{1}{c|}{98(15/0/83/2)} & \multicolumn{1}{c}{97(88/2/9/1)} & \multicolumn{1}{c}{96} & \multicolumn{1}{c}{\textcolor{blue}{99(20/0/79/1)}}\\
\multicolumn{1}{c}{81(79/19/2/0)} & \multicolumn{1}{c}{78(0/0/78/22)} & \multicolumn{1}{c|}{79(67/17/12/4)} & \multicolumn{1}{c}{78(78/19/0/3)} & \multicolumn{1}{c}{81} & \multicolumn{1}{c}{79(27/12/52/9)}\\
\multicolumn{1}{c}{73(37/12/36/15)} & \multicolumn{1}{c}{74(0/0/74/26)} & \multicolumn{1}{c|}{73(19/5/54/22)} & \multicolumn{1}{c}{70(59/27/11/3)} & \multicolumn{1}{c}{\textcolor{blue}{77}} & \multicolumn{1}{c}{72(3/2/69/26)}\\
\multicolumn{1}{c}{\textcolor{magenta}{91(82/8/9/1)}} & \multicolumn{1}{c}{89(0/0/89/11)} & \multicolumn{1}{c|}{89(49/6/40/5)} & \multicolumn{1}{c}{89(86/11/3/0)} & \multicolumn{1}{c}{86} & \multicolumn{1}{c}{90(32/4/58/6)}\\
\multicolumn{1}{c}{\textcolor{magenta}{\textbf{0.845}}} & \multicolumn{1}{c}{\textbf{0.794}} & \multicolumn{1}{c|}{\textbf{0.835}} & \multicolumn{1}{c}{\textbf{\textcolor{blue}{0.857}}} & \multicolumn{1}{c}{\textbf{\textcolor{blue}{0.882}}} & \multicolumn{1}{c}{\textbf{\textcolor{blue}{0.869}}}\\
\bottomrule
\end{tabular}
\end{sc}
\end{small}
\end{center}
\vskip -0.1in
\end{table*}

\newpage
\subsection{GPT-4o-mini}
\begin{table}[h]
\caption{Experimental results of GPT-4o-mini. Each item comprises the ratios of total success rate (code correct, code wrong, text correct, text wrong).}
\label{table: GPT-4o-mini-full-results-1}
\vskip 0.15in
\begin{center}
\begin{small}
\begin{sc}
\begin{tabular}{lcccccr}
\toprule
\multicolumn{1}{c}{} & \multicolumn{4}{c}{\textbf{Baseline Methods}}\\
\multicolumn{1}{c}{Task (success rate \%)} & \multicolumn{1}{c}{1.Only} & \multicolumn{1}{c}{2.All} & \multicolumn{1}{c}{3.All} & \multicolumn{1}{c}{4.All Code}\\
\multicolumn{1}{c}{} & \multicolumn{1}{c}{question} & \multicolumn{1}{c}{text} & \multicolumn{1}{c}{code} & \multicolumn{1}{c}{+ CoT}\\
\midrule
\multicolumn{1}{c}{Number Multi.} & \multicolumn{1}{c}{15(0/0/15/85)} & \multicolumn{1}{c}{26(0/0/26/74)} & \multicolumn{1}{c}{\textcolor{magenta}{100(100/0/0/0)}} & \multicolumn{1}{c}{\textcolor{magenta}{100(100/0/0/0)}}\\
\multicolumn{1}{c}{Game 24} & \multicolumn{1}{c}{15(0/0/15/85)} & \multicolumn{1}{c}{16(0/0/16/84)} & \multicolumn{1}{c}{9(9/91/0/0)} & \multicolumn{1}{c}{10(10/84/0/6)} \\
\multicolumn{1}{c}{Path plan} & \multicolumn{1}{c}{55(49/29/6/16)} & \multicolumn{1}{c}{21(0/0/21/79)} & \multicolumn{1}{c}{\textcolor{magenta}{58(58/42/0/0)}} & \multicolumn{1}{c}{49(49/51/0/0)} \\
\multicolumn{1}{c}{Letters} & \multicolumn{1}{c}{7(0/0/7/93)} & \multicolumn{1}{c}{78(0/0/78/22)} & \multicolumn{1}{c}{\textcolor{magenta}{100(100/0/0/0)}} & \multicolumn{1}{c}{\textcolor{magenta}{100(100/0/0/0)}} \\
\multicolumn{1}{c}{BoxLift}& \multicolumn{1}{c}{37.6} & \multicolumn{1}{c}{41.7} & \multicolumn{1}{c}{40.5} & \multicolumn{1}{c}{26.4} \\
\multicolumn{1}{c}{BoxNet} & \multicolumn{1}{c}{10.76} & \multicolumn{1}{c}{21.94} & \multicolumn{1}{c}{20.21} & \multicolumn{1}{c}{0} \\
\multicolumn{1}{c}{Blocksworld} & \multicolumn{1}{c}{17(0/0/17/83)} & \multicolumn{1}{c}{38(0/0/38/62)} & \multicolumn{1}{c}{17(17/83/0/0)} & \multicolumn{1}{c}{\textcolor{magenta}{40(40/60/0/0)}} \\
\multicolumn{1}{c}{Date Unde.} & \multicolumn{1}{c}{80(0/0/80/20)} & \multicolumn{1}{c}{\textcolor{magenta}{85(0/0/85/15)}} & \multicolumn{1}{c}{57(57/43/0/0)} & \multicolumn{1}{c}{70(70/30/0/0)} \\
\multicolumn{1}{c}{Web of Lies} & \multicolumn{1}{c}{\textcolor{magenta}{98(0/0/98/2)}} & \multicolumn{1}{c}{81(0/0/81/19)} & \multicolumn{1}{c}{70(70/30/0/0)} & \multicolumn{1}{c}{93(93/7/0/0)} \\
\multicolumn{1}{c}{Logical Dedu.} & \multicolumn{1}{c}{78(0/0/78/22)} & \multicolumn{1}{c}{\textcolor{magenta}{80(0/0/80/20)}} & \multicolumn{1}{c}{67(67/33/0/0)} & \multicolumn{1}{c}{73(73/27/0/0)} \\
\multicolumn{1}{c}{Navigate} & \multicolumn{1}{c}{\textcolor{magenta}{96(0/0/96/4)}} & \multicolumn{1}{c}{90(0/0/90/10)} & \multicolumn{1}{c}{89(87/9/2/2)} & \multicolumn{1}{c}{85(85/15/0/0)} \\
\multicolumn{1}{c}{GSM-Hard} & \multicolumn{1}{c}{73(0/0/73/27)} & \multicolumn{1}{c}{72(0/0/72/28)} & \multicolumn{1}{c}{77(77/23/0/0)} & \multicolumn{1}{c}{\textcolor{magenta}{80(80/20/0/0)}} \\
\multicolumn{1}{c}{MATH Geo.} & \multicolumn{1}{c}{73(0/0/73/27)} & \multicolumn{1}{c}{72(0/0/72/28)} & \multicolumn{1}{c}{72(72/28/0/0)} & \multicolumn{1}{c}{74(74/26/0/0)} \\
\multicolumn{1}{c}{MATH Count.$\And$Prob.} & \multicolumn{1}{c}{88(0/0/88/12)} & \multicolumn{1}{c}{\textcolor{magenta}{92(0/0/92/8)}} & \multicolumn{1}{c}{78(78/22/0/0)} & \multicolumn{1}{c}{83(83/17/0/0)} \\
\multicolumn{1}{c}{\textbf{Average Norm. Score}} & \multicolumn{1}{c}{\textbf{0.676}} & \multicolumn{1}{c}{\textbf{0.759}} & \multicolumn{1}{c}{\textbf{0.774}} & \multicolumn{1}{c}{\textbf{0.766}} \\
\bottomrule
\end{tabular}
\end{sc}
\end{small}
\end{center}
\vskip -0.1in
\end{table}

\begin{table*}[h]
%\caption{}
\label{table: GPT-4o-mini-full-results-2}
\vskip 0.15in
\begin{center}
\begin{small}
\begin{sc}
\begin{tabular}{lccccccr}
\toprule
\multicolumn{3}{c|}{\textbf{Baseline Methods}} & \multicolumn{3}{c}{\textbf{Proposed Methods}}\\
%\midrule
%\cline{2-5}
\multicolumn{1}{c}{5.AutoGen} & \multicolumn{1}{c}{6.AutoGen} & \multicolumn{1}{c|}{7.Code} & \multicolumn{1}{c}{8.Code} & \multicolumn{1}{c}{9.Code+t} & \multicolumn{1}{c}{10.Self-esti}\\
\multicolumn{1}{c}{conca.} & \multicolumn{1}{c}{system} & \multicolumn{1}{c|}{Interpreter} & \multicolumn{1}{c}{Interpreter+} & \multicolumn{1}{c}{ext+sum.} & \multicolumn{1}{c}{mate score}\\
\midrule
\multicolumn{1}{c}{1(0/98/1/1)} & \multicolumn{1}{c}{15(0/0/15/85)} & \multicolumn{1}{c|}{\textcolor{magenta}{100(100/0/0/0)}} & \multicolumn{1}{c}{\textcolor{blue}{100(100/0/0/0)}} & \multicolumn{1}{c}{99} & \multicolumn{1}{c}{42(31/2/11/56)}\\
\multicolumn{1}{c}{13(4/24/9/63)} & \multicolumn{1}{c}{14(1/4/13/82)} & \multicolumn{1}{c|}{\textcolor{magenta}{62(62/38/0/0)}} & \multicolumn{1}{c}{\textcolor{blue}{83(83/17/0/0)}} & \multicolumn{1}{c}{17} & \multicolumn{1}{c}{23(0/3/23/74)}\\
\multicolumn{1}{c}{51(51/49/0/0)} & \multicolumn{1}{c}{57(54/28/3/15)} & \multicolumn{1}{c|}{26(26/74/0/0)} & \multicolumn{1}{c}{26(26/74/0/0)} & \multicolumn{1}{c}{37} & \multicolumn{1}{c}{37(34/57/3/6)}\\
\multicolumn{1}{c}{\textcolor{magenta}{100(100/0/0/0)}} & \multicolumn{1}{c}{7(0/0/7/93)} & \multicolumn{1}{c|}{87(87/13/0/0)} & \multicolumn{1}{c}{89(89/11/0/0)} & \multicolumn{1}{c}{90} & \multicolumn{1}{c}{51(0/0/51/49)}\\
\multicolumn{1}{c}{37.4} & \multicolumn{1}{c}{39.4} & \multicolumn{1}{c|}{\textcolor{magenta}{44.7}} & \multicolumn{1}{c}{\textcolor{blue}{64.8}} & \multicolumn{1}{c}{42.8} & \multicolumn{1}{c}{38.2}\\
\multicolumn{1}{c}{16.87} & \multicolumn{1}{c}{13.18} & \multicolumn{1}{c|}{\textcolor{magenta}{23.78}} & \multicolumn{1}{c}{4.17} & \multicolumn{1}{c}{22.36} & \multicolumn{1}{c}{23.34}\\
\multicolumn{1}{c}{\textcolor{magenta}{40(0/0/40/60)}} & \multicolumn{1}{c}{15(0/0/15/85)} & \multicolumn{1}{c|}{17(3/20/14/63)} & \multicolumn{1}{c}{23(7/56/16/21)} & \multicolumn{1}{c}{38} & \multicolumn{1}{c}{34(0/0/34/66)}\\
\multicolumn{1}{c}{63(49/35/14/2)} & \multicolumn{1}{c}{80(0/0/80/20)} & \multicolumn{1}{c|}{74(74/26/0/0)} & \multicolumn{1}{c}{77(73/23/4/0)} & \multicolumn{1}{c}{83} & \multicolumn{1}{c}{82(0/0/82/18)}\\
\multicolumn{1}{c}{76(0/0/76/24)} & \multicolumn{1}{c}{96(0/0/96/4)} & \multicolumn{1}{c|}{59(42/38/17/3)} & \multicolumn{1}{c}{52(30/33/22/15)} & \multicolumn{1}{c}{82} & \multicolumn{1}{c}{83(0/0/83/17)}\\
\multicolumn{1}{c}{75(2/0/73/25)} & \multicolumn{1}{c}{76(0/0/76/24)} & \multicolumn{1}{c|}{75(75/25/0/0)} & \multicolumn{1}{c}{78(56/17/22/5)} & \multicolumn{1}{c}{\textcolor{blue}{82}} & \multicolumn{1}{c}{73(0/0/73/27)}\\
\multicolumn{1}{c}{55(0/0/55/45)} & \multicolumn{1}{c}{95(0/0/95/5)} & \multicolumn{1}{c|}{94(94/6/0/0)} & \multicolumn{1}{c}{\textcolor{blue}{96(84/4/12/0)}} & \multicolumn{1}{c}{95} & \multicolumn{1}{c}{94(0/0/94/6)}\\
\multicolumn{1}{c}{68(65/32/3/0)} & \multicolumn{1}{c}{73(0/0/73/27)} & \multicolumn{1}{c|}{73(0/0/73/27)} & \multicolumn{1}{c}{52(52/48/0/0)} & \multicolumn{1}{c}{77} & \multicolumn{1}{c}{73(0/0/73/27)}\\
\multicolumn{1}{c}{74(29/6/45/20)} & \multicolumn{1}{c}{76(0/0/76/24)} & \multicolumn{1}{c|}{\textcolor{magenta}{77(0/0/77/23)}} & \multicolumn{1}{c}{\textcolor{blue}{81(74/17/7/2)}} & \multicolumn{1}{c}{72} & \multicolumn{1}{c}{74(0/0/74/26)}\\
\multicolumn{1}{c}{88(66/9/22/3)} & \multicolumn{1}{c}{88(0/0/88/12)} & \multicolumn{1}{c|}{83(0/0/83/17)} & \multicolumn{1}{c}{87(85/12/2/1)} & \multicolumn{1}{c}{88} & \multicolumn{1}{c}{87(1/0/86/13)}\\
\multicolumn{1}{c}{\textbf{0.718}} & \multicolumn{1}{c}{\textbf{0.682}} & \multicolumn{1}{c|}{\textbf{\textcolor{magenta}{0.808}}} & \multicolumn{1}{c}{\textbf{0.790}} & \multicolumn{1}{c}{\textbf{\textcolor{blue}{0.850}}} & \multicolumn{1}{c}{\textbf{0.765}}\\
\bottomrule
\end{tabular}
\end{sc}
\end{small}
\end{center}
\vskip -0.1in
\end{table*}

\newpage
\subsection{GPT-3.5}
\begin{table}[h]
\caption{Experimental results of GPT-3.5-turbo. Each item comprises the ratios of total success rate (code correct, code wrong, text correct, text wrong).}
\label{table: GPT-3.5-turbo-full-results-1}
\vskip 0.15in
\begin{center}
\begin{small}
\begin{sc}
\begin{tabular}{lcccccr}
\toprule
\multicolumn{1}{c}{} & \multicolumn{4}{c}{\textbf{Baseline Methods}}\\
\multicolumn{1}{c}{Task (success rate \%)} & \multicolumn{1}{c}{1.Only} & \multicolumn{1}{c}{2.All} & \multicolumn{1}{c}{3.All} & \multicolumn{1}{c}{4.All Code}\\
\multicolumn{1}{c}{} & \multicolumn{1}{c}{question} & \multicolumn{1}{c}{text} & \multicolumn{1}{c}{code} & \multicolumn{1}{c}{+ CoT}\\
\midrule
\multicolumn{1}{c}{Number Multi.} & \multicolumn{1}{c}{2(0/0/2/98)} & \multicolumn{1}{c}{15(0/0/15/85)} & \multicolumn{1}{c}{\textcolor{magenta}{100(100/0/0/0)}} & \multicolumn{1}{c}{\textcolor{magenta}{100(100/0/0/0)}}\\
\multicolumn{1}{c}{Game 24} & \multicolumn{1}{c}{3(0/0/3/97)} & \multicolumn{1}{c}{4(0/0/4/96)} & \multicolumn{1}{c}{12(12/88/0/0)} & \multicolumn{1}{c}{9(9/91/0/0)} \\
\multicolumn{1}{c}{Path plan} & \multicolumn{1}{c}{5(1/5/4/90)} & \multicolumn{1}{c}{8(0/0/8/92)} & \multicolumn{1}{c}{\textcolor{magenta}{36(36/64/0/0)}} & \multicolumn{1}{c}{16(16/75/0/9)} \\
\multicolumn{1}{c}{Letters} & \multicolumn{1}{c}{4(0/0/4/96)} & \multicolumn{1}{c}{44(0/0/44/56)} & \multicolumn{1}{c}{\textcolor{magenta}{100(100/0/0/0)}} & \multicolumn{1}{c}{\textcolor{magenta}{100(100/0/0/0)}} \\
\multicolumn{1}{c}{BoxLift}& \multicolumn{1}{c}{\textcolor{magenta}{37.8}} & \multicolumn{1}{c}{25.2} & \multicolumn{1}{c}{21.1} & \multicolumn{1}{c}{5.3} \\
\multicolumn{1}{c}{BoxNet} & \multicolumn{1}{c}{6.01} & \multicolumn{1}{c}{\textcolor{magenta}{7.35}} & \multicolumn{1}{c}{0} & \multicolumn{1}{c}{0} \\
\multicolumn{1}{c}{Blocksworld} & \multicolumn{1}{c}{6(0/0/6/94)} & \multicolumn{1}{c}{\textcolor{magenta}{13(0/0/13/87)}} & \multicolumn{1}{c}{0(0/100/0/0)} & \multicolumn{1}{c}{3(1/67/2/30)} \\
\multicolumn{1}{c}{Date Unde.} & \multicolumn{1}{c}{61(0/0/61/39)} & \multicolumn{1}{c}{53(0/0/53/47)} & \multicolumn{1}{c}{50(50/50/0/0)} & \multicolumn{1}{c}{38(38/62/0/0)} \\
\multicolumn{1}{c}{Web of Lies} & \multicolumn{1}{c}{53(0/0/53/47)} & \multicolumn{1}{c}{67(0/0/67/33)} & \multicolumn{1}{c}{\textcolor{magenta}{68(68/32/0/0)}} & \multicolumn{1}{c}{63(56/28/7/9)} \\
\multicolumn{1}{c}{Logical Dedu.} & \multicolumn{1}{c}{32(0/0/32/68)} & \multicolumn{1}{c}{\textcolor{magenta}{38(0/0/38/62)}} & \multicolumn{1}{c}{\textcolor{magenta}{38(38/62/0/0)}} & \multicolumn{1}{c}{27(27/72/0/1)} \\
\multicolumn{1}{c}{Navigate} & \multicolumn{1}{c}{52(0/0/52/48)} & \multicolumn{1}{c}{78(0/0/78/22)} & \multicolumn{1}{c}{75(75/25/0/0)} & \multicolumn{1}{c}{\textcolor{magenta}{90(89/9/1/1)}} \\
\multicolumn{1}{c}{GSM-Hard} & \multicolumn{1}{c}{60(0/0/60/40)} & \multicolumn{1}{c}{56(0/0/56/44)} & \multicolumn{1}{c}{\textcolor{magenta}{67(67/33/0/0)}} & \multicolumn{1}{c}{63(61/33/2/4)} \\
\multicolumn{1}{c}{MATH Geo.} & \multicolumn{1}{c}{57(0/0/57/43)} & \multicolumn{1}{c}{48(0/0/48/52)} & \multicolumn{1}{c}{52(45/46/7/2)} & \multicolumn{1}{c}{57(41/38/16/5)} \\
\multicolumn{1}{c}{MATH Count.$\And$Prob.} & \multicolumn{1}{c}{59(0/0/59/41)} & \multicolumn{1}{c}{58(0/0/58/42)} & \multicolumn{1}{c}{71(65/27/6/2)} & \multicolumn{1}{c}{\textcolor{magenta}{77(60/18/17/5)}} \\
\multicolumn{1}{c}{\textbf{Average Norm. Score}} & \multicolumn{1}{c}{\textbf{0.653}} & \multicolumn{1}{c}{\textbf{0.653}} & \multicolumn{1}{c}{\textbf{\textcolor{magenta}{0.6813}}} & \multicolumn{1}{c}{\textbf{0.640}} \\
\bottomrule
\end{tabular}
\end{sc}
\end{small}
\end{center}
\vskip -0.1in
\end{table}

\begin{table*}[h]
%\caption{}
\label{table: GPT-3.5-turbo-full-results-2}
\vskip 0.15in
\begin{center}
\begin{small}
\begin{sc}
\begin{tabular}{lccccccr}
\toprule
\multicolumn{3}{c|}{\textbf{Baseline Methods}} & \multicolumn{3}{c}{\textbf{Proposed Methods}}\\
%\midrule
%\cline{2-5}
\multicolumn{1}{c}{5.AutoGen} & \multicolumn{1}{c}{6.AutoGen} & \multicolumn{1}{c|}{7.Code} & \multicolumn{1}{c}{8.Code} & \multicolumn{1}{c}{9.Code+t} & \multicolumn{1}{c}{10.Self-esti}\\
\multicolumn{1}{c}{conca.} & \multicolumn{1}{c}{system} & \multicolumn{1}{c|}{Interpreter} & \multicolumn{1}{c}{Interpreter+} & \multicolumn{1}{c}{ext+sum.} & \multicolumn{1}{c}{mate score}\\
\midrule
\multicolumn{1}{c}{50(48/0/2/50)} & \multicolumn{1}{c}{2(0/0/2/98)} & \multicolumn{1}{c|}{\textcolor{magenta}{100(100/0/0/0)}} & \multicolumn{1}{c}{\textcolor{blue}{100(100/0/0/0)}} & \multicolumn{1}{c}{26} & \multicolumn{1}{c}{24(15/0/9/76)}\\
\multicolumn{1}{c}{91(91/9/0/0)} & \multicolumn{1}{c}{4(0/0/4/96)} & \multicolumn{1}{c|}{\textcolor{magenta}{92(92/8/0/0)}} & \multicolumn{1}{c}{6(6/94/0/0)} & \multicolumn{1}{c}{3} & \multicolumn{1}{c}{10(1/2/9/88)}\\
\multicolumn{1}{c}{1(1/9/0/90)} & \multicolumn{1}{c}{3(0/2/3/95)} & \multicolumn{1}{c|}{0(0/64/0/36)} & \multicolumn{1}{c}{0(0/76/0/24)} & \multicolumn{1}{c}{23} & \multicolumn{1}{c}{12(6/6/6/82)}\\
\multicolumn{1}{c}{\textcolor{magenta}{100(100/0/0/0)}} & \multicolumn{1}{c}{4(0/0/4/96)} & \multicolumn{1}{c|}{70(70/21/0/9)} & \multicolumn{1}{c}{89(89/7/0/4)} & \multicolumn{1}{c}{67} & \multicolumn{1}{c}{5(0/0/5/95)}\\
\multicolumn{1}{c}{12.2} & \multicolumn{1}{c}{37.4} & \multicolumn{1}{c|}{3.5} & \multicolumn{1}{c}{4.7} & \multicolumn{1}{c}{28.6} & \multicolumn{1}{c}{16.3}\\
\multicolumn{1}{c}{1.39} & \multicolumn{1}{c}{6.17} & \multicolumn{1}{c|}{0} & \multicolumn{1}{c}{0} & \multicolumn{1}{c}{4.93} & \multicolumn{1}{c}{\textcolor{blue}{12.34}}\\
\multicolumn{1}{c}{9(0/0/9/91)} & \multicolumn{1}{c}{3(0/0/3/97)} & \multicolumn{1}{c|}{12(7/63/5/25)} & \multicolumn{1}{c}{7(7/93/0/0)} & \multicolumn{1}{c}{3} & \multicolumn{1}{c}{8(0/0/8/92)}\\
\multicolumn{1}{c}{20(20/78/0/2)} & \multicolumn{1}{c}{\textcolor{magenta}{62(0/0/62/38)}} & \multicolumn{1}{c|}{25(20/37/5/38)} & \multicolumn{1}{c}{36(35/53/1/11)} & \multicolumn{1}{c}{\textcolor{blue}{62}} & \multicolumn{1}{c}{56(0/0/56/44)}\\
\multicolumn{1}{c}{53(2/2/51/45)} & \multicolumn{1}{c}{55(0/0/55/45)} & \multicolumn{1}{c|}{42(8/5/34/53)} & \multicolumn{1}{c}{52(48/38/4/10)} & \multicolumn{1}{c}{56} & \multicolumn{1}{c}{40(0/0/40/60)}\\
\multicolumn{1}{c}{31(14/48/17/21)} & \multicolumn{1}{c}{36(0/0/36/64)} & \multicolumn{1}{c|}{32(2/3/30/65)} & \multicolumn{1}{c}{28(26/60/2/12)} & \multicolumn{1}{c}{36} & \multicolumn{1}{c}{37(0/0/37/63)}\\
\multicolumn{1}{c}{60(0/0/60/40)} & \multicolumn{1}{c}{54(0/0/54/46)} & \multicolumn{1}{c|}{67(55/22/12/11)} & \multicolumn{1}{c}{81(71/11/10/8)} & \multicolumn{1}{c}{78} & \multicolumn{1}{c}{64(0/0/64/36)}\\
\multicolumn{1}{c}{55(51/30/4/15)} & \multicolumn{1}{c}{59(0/0/59/41)} & \multicolumn{1}{c|}{64(62/35/2/1)} & \multicolumn{1}{c}{63(63/31/0/6)} & \multicolumn{1}{c}{62} & \multicolumn{1}{c}{60(1/1/59/39)}\\
\multicolumn{1}{c}{\textcolor{magenta}{66(31/8/35/26)}} & \multicolumn{1}{c}{56(0/0/56/44)} & \multicolumn{1}{c|}{59(21/26/38/15)} & \multicolumn{1}{c}{53(29/35/24/12)} & \multicolumn{1}{c}{43} & \multicolumn{1}{c}{44(0/0/44/56)}\\
\multicolumn{1}{c}{76(46/8/30/16)} & \multicolumn{1}{c}{60(0/0/60/40)} & \multicolumn{1}{c|}{66(38/23/28/11)} & \multicolumn{1}{c}{66(32/23/34/11)} & \multicolumn{1}{c}{57} & \multicolumn{1}{c}{60(4/0/56/40)}\\
\multicolumn{1}{c}{\textbf{0.646}} & \multicolumn{1}{c}{\textbf{0.555}} & \multicolumn{1}{c|}{\textbf{0.645}} & \multicolumn{1}{c}{\textbf{0.585}} & \multicolumn{1}{c}{\textbf{0.639}} & \multicolumn{1}{c}{\textbf{0.592}}\\
\bottomrule
\end{tabular}
\end{sc}
\end{small}
\end{center}
\vskip -0.1in
\end{table*}

\newpage
\subsection{O1-preview}
\begin{table}[h]
\caption{Experimental results of O1-preview. Each item comprises the ratios of total success rate (code correct, code wrong, text correct, text wrong).}
\label{table: O1-preview-full-results-1}
\vskip 0.15in
\begin{center}
\begin{small}
\begin{sc}
\begin{tabular}{lcccccr}
\toprule
\multicolumn{1}{c}{} & \multicolumn{4}{c}{\textbf{Baseline Methods}}\\
\multicolumn{1}{c}{Task (success rate \%)} & \multicolumn{1}{c}{1.Only} & \multicolumn{1}{c}{2.All} & \multicolumn{1}{c}{3.All} & \multicolumn{1}{c}{4.All Code}\\
\multicolumn{1}{c}{} & \multicolumn{1}{c}{question} & \multicolumn{1}{c}{text} & \multicolumn{1}{c}{code} & \multicolumn{1}{c}{+ CoT}\\
\midrule
\multicolumn{1}{c}{Game 24} & \multicolumn{1}{c}{78(0,0,78,22)} & \multicolumn{1}{c}{69(0,0,69,31)} & \multicolumn{1}{c}{82(58,12,24,6)} & \multicolumn{1}{c}{\textcolor{magenta}{87(67,8,20,5)}} \\
\multicolumn{1}{c}{Path plan} & 56(52,37,4,7) & 61(3,3,58,36) & 59(59,41,0,0) & \textcolor{magenta}{64(64,36,0,0)} \\
\multicolumn{1}{c}{BoxLift} & 67.09 & 56.24 & 85.88 & \textcolor{magenta}{91.57} \\
\multicolumn{1}{c}{BoxNet} & \textcolor{magenta}{67.34} & 59.92 & 63.68 & 49.55 \\
\multicolumn{1}{c}{Blocksworld} & 77(0,0,77,23) & 72(0,0,72,28) & 78(27,9,51,13) & 77(31,8,46,15) \\
\multicolumn{1}{c}{\textbf{Average Norm. Score}} & \multicolumn{1}{c}{\textbf{0.8820}} & \multicolumn{1}{c}{\textbf{0.8195}} & \multicolumn{1}{c}{\textbf{\textcolor{magenta}{0.9331}}} & \multicolumn{1}{c}{\textbf{0.9283}} \\
\bottomrule
\end{tabular}
\end{sc}
\end{small}
\end{center}
\vskip -0.1in
\end{table}

\begin{table*}[h]
%\caption{Scores on various tasks for different methods. Total success rate (code correct, code wrong, text correct, text wrong) \%.}
\label{table: O1-preview-full-results-2}
\vskip 0.15in
\begin{center}
\begin{small}
\begin{sc}
\begin{tabular}{lccc}
\toprule
\multicolumn{1}{c|}{\textbf{Baseline Methods}} & \multicolumn{2}{c}{\textbf{Proposed Methods}}\\
\multicolumn{1}{c|}{5.AutoGen Conca.} & \multicolumn{1}{c}{9.Code+text+sum.} & \multicolumn{1}{c}{10.Self-estimate Score}\\
\midrule
\multicolumn{1}{c|}{69(1,7,68,24)} & 77 & 63(1,3,62,34) \\
\multicolumn{1}{c|}{56(25,26,31,18)} & 61 & 47(24,33,23,20) \\
\multicolumn{1}{c|}{73.92} & 72.08 & 38.48 \\
\multicolumn{1}{c|}{48.59} & 62.81 & 64.34 \\
\multicolumn{1}{c|}{\textcolor{magenta}{85(0,0,85,15)}} & 81 & 79(0,1,79,20) \\
\multicolumn{1}{c|}{\textbf{0.8394}} & \multicolumn{1}{c}{\textbf{0.9022}} & \multicolumn{1}{c}{\textbf{0.7527}} \\
\bottomrule
\end{tabular}
\end{sc}
\end{small}
\end{center}
\vskip -0.1in
\end{table*}

\newpage
\subsection{Claude-3-sonnet-20240229}
\begin{table}[h]
\caption{Experimental results of Claude-3-sonnet-20240229. Each item comprises the ratios of total success rate (code correct, code wrong, text correct, text wrong).}
\label{table: Claude-sonnet-full-results-1}
\vskip 0.15in
\begin{center}
\begin{small}
\begin{sc}
\begin{tabular}{lcccccr}
\toprule
\multicolumn{1}{c}{} & \multicolumn{4}{c}{\textbf{Baseline Methods}}\\
\multicolumn{1}{c}{Task (success rate \%)} & \multicolumn{1}{c}{1.Only} & \multicolumn{1}{c}{2.All} & \multicolumn{1}{c}{3.All} & \multicolumn{1}{c}{4.All Code}\\
\multicolumn{1}{c}{} & \multicolumn{1}{c}{question} & \multicolumn{1}{c}{text} & \multicolumn{1}{c}{code} & \multicolumn{1}{c}{+ CoT}\\
\midrule
\multicolumn{1}{c}{Number Multi.} & \multicolumn{1}{c}{37(0/0/37/63)} & \multicolumn{1}{c}{32(0/0/32/68)} & \multicolumn{1}{c}{\textcolor{magenta}{100(100/0/0/0)}} & \multicolumn{1}{c}{99(99/1/0/0)}\\
\multicolumn{1}{c}{Game 24} & \multicolumn{1}{c}{4(0/0/4/96)} & \multicolumn{1}{c}{4(0/0/4/96)} & \multicolumn{1}{c}{4(4/96/0/0)} & \multicolumn{1}{c}{\textcolor{magenta}{34(34/66/0/0)}} \\
\multicolumn{1}{c}{Path plan} & \multicolumn{1}{c}{33(4/6/29/61)} & \multicolumn{1}{c}{26(0/0/26/74)} & \multicolumn{1}{c}{\textcolor{magenta}{58(58/42/0/0)}} & \multicolumn{1}{c}{42(42/58/0/0)} \\
\multicolumn{1}{c}{Letters} & \multicolumn{1}{c}{2(0/0/2/98)} & \multicolumn{1}{c}{9(0/0/9/91)} & \multicolumn{1}{c}{\textcolor{magenta}{100(100/0/0/0)}} & \multicolumn{1}{c}{\textcolor{magenta}{100(100/0/0/0)}} \\
\multicolumn{1}{c}{BoxLift}& \multicolumn{1}{c}{48.02} & \multicolumn{1}{c}{\textcolor{magenta}{50.31}} & \multicolumn{1}{c}{12.23} & \multicolumn{1}{c}{8.19} \\
\multicolumn{1}{c}{BoxNet} & \multicolumn{1}{c}{21.36} & \multicolumn{1}{c}{21.01} & \multicolumn{1}{c}{24.26} & \multicolumn{1}{c}{\textcolor{magenta}{30.50}} \\
\multicolumn{1}{c}{Blocksworld} & \multicolumn{1}{c}{23(0/0/23/77)} & \multicolumn{1}{c}{\textcolor{magenta}{26(0/0/26/74)}} & \multicolumn{1}{c}{6(0/14/6/80)} & \multicolumn{1}{c}{21(3/3/19/75)} \\
\multicolumn{1}{c}{Date Unde.} & \multicolumn{1}{c}{\textcolor{magenta}{73(0/0/73/27)}} & \multicolumn{1}{c}{71(0/0/71/29)} & \multicolumn{1}{c}{54(54/46/0/0)} & \multicolumn{1}{c}{63(63/37/0/0)} \\
\multicolumn{1}{c}{Web of Lies} & \multicolumn{1}{c}{88(0/0/88/12)} & \multicolumn{1}{c}{\textcolor{magenta}{89(0/0/89/11)}} & \multicolumn{1}{c}{86(86/14/0/0)} & \multicolumn{1}{c}{59(58/40/2/0)} \\
\multicolumn{1}{c}{Logical Dedu.} & \multicolumn{1}{c}{60(0/0/60/40)} & \multicolumn{1}{c}{\textcolor{magenta}{61(0/0/61/39)}} & \multicolumn{1}{c}{42(42/58/0/0)} & \multicolumn{1}{c}{53(53/47/0/0)} \\
\multicolumn{1}{c}{Navigate} & \multicolumn{1}{c}{76(0/0/76/24)} & \multicolumn{1}{c}{79(0/0/79/21)} & \multicolumn{1}{c}{89(89/11/0/0)} & \multicolumn{1}{c}{\textcolor{magenta}{94(94/6/0/0)}} \\
\multicolumn{1}{c}{GSM} & \multicolumn{1}{c}{70(0/0/70/30)} & \multicolumn{1}{c}{72(0/0/72/28)} & \multicolumn{1}{c}{76(76/24/0/0)} & \multicolumn{1}{c}{\textcolor{magenta}{77(77/23/0/0)}} \\
\multicolumn{1}{c}{MATH Geo.} & \multicolumn{1}{c}{44(0/0/44/56)} & \multicolumn{1}{c}{44(0/0/44/56)} & \multicolumn{1}{c}{39(39/61/0/0)} & \multicolumn{1}{c}{37(37/63/0/0)} \\
\multicolumn{1}{c}{MATH Count.$\&$Prob.} & \multicolumn{1}{c}{65(2/0/63/35)} & \multicolumn{1}{c}{63(0/0/63/37)} & \multicolumn{1}{c}{76(76/24/0/0)} & \multicolumn{1}{c}{\textcolor{magenta}{77(77/23/0/0)}} \\
\multicolumn{1}{c}{\textbf{Average Norm. Score}} & \multicolumn{1}{c}{\textbf{0.715}} & \multicolumn{1}{c}{\textbf{0.720}} & \multicolumn{1}{c}{\textbf{0.747}} & \multicolumn{1}{c}{\textbf{\textcolor{magenta}{0.810}}} \\
\bottomrule
\end{tabular}
\end{sc}
\end{small}
\end{center}
\vskip -0.1in
\end{table}

\begin{table*}[h]
%\caption{}
\label{table: Claude-sonnet-full-results-2}
\vskip 0.15in
\begin{center}
\begin{small}
\begin{sc}
\begin{tabular}{lccccr}
\toprule
\multicolumn{2}{c|}{\textbf{Baseline Methods}} & \multicolumn{2}{c}{\textbf{Proposed Methods}}\\
%\midrule
%\cline{2-5}
\multicolumn{1}{c}{5.AutoGen} & \multicolumn{1}{c|}{6.AutoGen} & \multicolumn{1}{c}{9.Code+t} & \multicolumn{1}{c}{10.Self-esti}\\
\multicolumn{1}{c}{conca.} & \multicolumn{1}{c|}{system} & \multicolumn{1}{c}{ext+sum.} & \multicolumn{1}{c}{mate score}\\
\midrule
\multicolumn{1}{c}{\textcolor{magenta}{100(100/0/0/0)}} & \multicolumn{1}{c|}{37(0/0/37/63)} & \multicolumn{1}{c}{95} & \multicolumn{1}{c}{87(82/0/5/13)}\\
\multicolumn{1}{c}{5(5/95/0/0)} & \multicolumn{1}{c|}{4(0/0/4/96)} & \multicolumn{1}{c}{7} & \multicolumn{1}{c}{\textcolor{blue}{69(69/31/0/0)}}\\
\multicolumn{1}{c}{56(56/44/0/0)} & \multicolumn{1}{c|}{33(4/6/29/61)} & \multicolumn{1}{c}{54} & \multicolumn{1}{c}{41(41/59/0/0)}\\
\multicolumn{1}{c}{99(99/0/0/1)} & \multicolumn{1}{c|}{2(0/0/2/98)} & \multicolumn{1}{c}{74} & \multicolumn{1}{c}{34(26/0/8/66)}\\
\multicolumn{1}{c}{5.05} & \multicolumn{1}{c|}{48.02} & \multicolumn{1}{c}{28.79} & \multicolumn{1}{c}{0.51}\\
\multicolumn{1}{c}{28.82} & \multicolumn{1}{c|}{21.36} & \multicolumn{1}{c}{19.95} & \multicolumn{1}{c}{1.89}\\
\multicolumn{1}{c}{7(6/81/1/12)} & \multicolumn{1}{c|}{23(0/0/23/77)} & \multicolumn{1}{c}{19} & \multicolumn{1}{c}{15(0/54/15/31)}\\
\multicolumn{1}{c}{60(56/34/4/6)} & \multicolumn{1}{c|}{\textcolor{magenta}{73(0/0/73/27)}} & \multicolumn{1}{c}{72} & \multicolumn{1}{c}{68(26/19/42/13)}\\
\multicolumn{1}{c}{58(48/35/10/7)} & \multicolumn{1}{c|}{86(0/0/86/14)} & \multicolumn{1}{c}{54} & \multicolumn{1}{c}{82(0/0/82/18)}\\
\multicolumn{1}{c}{55(55/45/0/0)} & \multicolumn{1}{c|}{\textcolor{magenta}{62(0/0/62/38)}} & \multicolumn{1}{c}{61} & \multicolumn{1}{c}{60(2/4/58/36)}\\
\multicolumn{1}{c}{74(62/21/12/5)} & \multicolumn{1}{c|}{76(0/0/76/24)} & \multicolumn{1}{c}{80} & \multicolumn{1}{c}{78(0/0/78/22)}\\
\multicolumn{1}{c}{68(63/29/5/3)} & \multicolumn{1}{c|}{70(0/0/70/30)} & \multicolumn{1}{c}{69} & \multicolumn{1}{c}{62(18/25/44/13)}\\
\multicolumn{1}{c}{\textcolor{magenta}{47(43/47/4/6)}} & \multicolumn{1}{c|}{42(0/0/42/58)} & \multicolumn{1}{c}{39(9/11/30/50)} & \multicolumn{1}{c}{42(21/36/21/22)}\\
\multicolumn{1}{c}{\textcolor{magenta}{77(72/20/5/3)}} & \multicolumn{1}{c|}{64(2/0/62/36)} & \multicolumn{1}{c}{65} & \multicolumn{1}{c}{62(24/11/38/27)}\\
\multicolumn{1}{c}{\textbf{0.740}} & \multicolumn{1}{c|}{\textbf{0.711}} & \multicolumn{1}{c}{\textbf{0.762}} & \multicolumn{1}{c}{\textbf{0.694}}\\
\bottomrule
\end{tabular}
\end{sc}
\end{small}
\end{center}
\vskip -0.1in
\end{table*}

\newpage
\subsection{Open-mixtral-8x7b}
\begin{table}[h]
\caption{Experimental results of Open-mixtral-8x7b. Each item comprises the ratios of total success rate (code correct, code wrong, text correct, text wrong).}
\label{table: Open-mixtral-8x7b-full-results-1}
\vskip 0.15in
\begin{center}
\begin{small}
\begin{sc}
\begin{tabular}{lcccccr}
\toprule
\multicolumn{1}{c}{} & \multicolumn{4}{c}{\textbf{Baseline Methods}}\\
\multicolumn{1}{c}{Task (success rate \%)} & \multicolumn{1}{c}{1.Only} & \multicolumn{1}{c}{2.All} & \multicolumn{1}{c}{3.All} & \multicolumn{1}{c}{4.All Code}\\
\multicolumn{1}{c}{} & \multicolumn{1}{c}{question} & \multicolumn{1}{c}{text} & \multicolumn{1}{c}{code} & \multicolumn{1}{c}{+ CoT}\\
\midrule
\multicolumn{1}{c}{Number Multi.} & \multicolumn{1}{c}{5(0/0/5/95)} & \multicolumn{1}{c}{5(0/0/5/95)} & \multicolumn{1}{c}{\textcolor{magenta}{100(100/0/0/0)}} & \multicolumn{1}{c}{94(94/5/0/1)}\\
\multicolumn{1}{c}{Game 24} & \multicolumn{1}{c}{\textcolor{magenta}{4(0/0/4/96)}} & \multicolumn{1}{c}{2(0/0/2/98)} & \multicolumn{1}{c}{1(1/99/0/0)} & \multicolumn{1}{c}{2(2/94/0/4)} \\
\multicolumn{1}{c}{Path plan} & \multicolumn{1}{c}{5(0/1/5/94)} & \multicolumn{1}{c}{11(0/0/11/89)} & \multicolumn{1}{c}{16(16/84/0/0)} & \multicolumn{1}{c}{8(8/92/0/0)} \\
\multicolumn{1}{c}{Letters} & \multicolumn{1}{c}{1(0/0/1/99)} & \multicolumn{1}{c}{3(0/0/3/97)} & \multicolumn{1}{c}{82(82/18/0/0)} & \multicolumn{1}{c}{\textcolor{magenta}{95(95/5/0/0)}} \\
\multicolumn{1}{c}{BoxLift}& \multicolumn{1}{c}{\textcolor{magenta}{28.13}} & \multicolumn{1}{c}{22.69} & \multicolumn{1}{c}{2.32} & \multicolumn{1}{c}{4.83} \\
\multicolumn{1}{c}{BoxNet} & \multicolumn{1}{c}{5.15} & \multicolumn{1}{c}{\textcolor{magenta}{5.99}} & \multicolumn{1}{c}{3.13} & \multicolumn{1}{c}{0} \\
\multicolumn{1}{c}{Blocksworld} & \multicolumn{1}{c}{3(0/0/3/97)} & \multicolumn{1}{c}{\textcolor{magenta}{8(0/0/8/92)}} & \multicolumn{1}{c}{0(0/100/0/0)} & \multicolumn{1}{c}{0(0/100/0/0)} \\
\multicolumn{1}{c}{Date Unde.} & \multicolumn{1}{c}{48(0/0/48/52)} & \multicolumn{1}{c}{45(0/0/45/55)} & \multicolumn{1}{c}{50(50/50/0/0)} & \multicolumn{1}{c}{\textcolor{magenta}{56(56/40/0/4)}} \\
\multicolumn{1}{c}{Web of Lies} & \multicolumn{1}{c}{53(0/0/53/47)} & \multicolumn{1}{c}{\textcolor{magenta}{66(0/0/66/34)}} & \multicolumn{1}{c}{56(56/42/0/2)} & \multicolumn{1}{c}{61(61/38/0/1)} \\
\multicolumn{1}{c}{Logical Dedu.} & \multicolumn{1}{c}{36(0/0/36/64)} & \multicolumn{1}{c}{\textcolor{magenta}{41(0/0/41/59)}} & \multicolumn{1}{c}{32(32/68/0/0)} & \multicolumn{1}{c}{34(32/61/2/5)} \\
\multicolumn{1}{c}{Navigate} & \multicolumn{1}{c}{58(0/0/58/42)} & \multicolumn{1}{c}{41(0/0/41/59)} & \multicolumn{1}{c}{58(57/42/1/0)} & \multicolumn{1}{c}{\textcolor{magenta}{69(65/30/4/1)}} \\
\multicolumn{1}{c}{GSM} & \multicolumn{1}{c}{50(0/0/50/50)} & \multicolumn{1}{c}{48(0/0/48/52)} & \multicolumn{1}{c}{60(60/40/0/0)} & \multicolumn{1}{c}{\textcolor{magenta}{62(61/33/1/5)}} \\
\multicolumn{1}{c}{MATH Geo.} & \multicolumn{1}{c}{42(0/0/42/58)} & \multicolumn{1}{c}{44(0/0/44/56)} & \multicolumn{1}{c}{46(44/49/2/5)} & \multicolumn{1}{c}{\textcolor{magenta}{49(47/45/2/6)}} \\
\multicolumn{1}{c}{MATH Count.$\&$Prob.} & \multicolumn{1}{c}{50(0/0/50/50)} & \multicolumn{1}{c}{54(0/0/54/46)} & \multicolumn{1}{c}{\textcolor{magenta}{74(73/25/1/1)}} & \multicolumn{1}{c}{64(60/35/4/1)} \\
\multicolumn{1}{c}{\textbf{Average Norm. Score}} & \multicolumn{1}{c}{\textbf{0.6345}} & \multicolumn{1}{c}{\textbf{0.6809}} & \multicolumn{1}{c}{\textbf{0.6923}} & \multicolumn{1}{c}{\textbf{0.6734}} \\
\bottomrule
\end{tabular}
\end{sc}
\end{small}
\end{center}
\vskip -0.1in
\end{table}

\begin{table*}[h]
%\caption{}
\label{table: Open-mixtral-8x7b-full-results-2}
\vskip 0.15in
\begin{center}
\begin{small}
\begin{sc}
\begin{tabular}{lccccr}
\toprule
\multicolumn{2}{c|}{\textbf{Baseline Methods}} & \multicolumn{2}{c}{\textbf{Proposed Methods}}\\
%\midrule
%\cline{2-5}
\multicolumn{1}{c}{5.AutoGen} & \multicolumn{1}{c|}{6.AutoGen} & \multicolumn{1}{c}{9.Code+t} & \multicolumn{1}{c}{10.Self-esti}\\
\multicolumn{1}{c}{conca.} & \multicolumn{1}{c|}{system} & \multicolumn{1}{c}{ext+sum.} & \multicolumn{1}{c}{mate score}\\
\midrule
\multicolumn{1}{c}{93(93/5/0/2)} & \multicolumn{1}{c|}{5(0/0/5/95)} & \multicolumn{1}{c}{68} & \multicolumn{1}{c}{59(51/38/8/3)}\\
\multicolumn{1}{c}{2(1/80/1/18)} & \multicolumn{1}{c|}{\textcolor{magenta}{4(0/0/4/96)}} & \multicolumn{1}{c}{\textcolor{blue}{6}} & \multicolumn{1}{c}{2(2/99/0/0)}\\
\multicolumn{1}{c}{\textcolor{magenta}{19(19/76/0/5)}} & \multicolumn{1}{c|}{5(0/1/5/94)} & \multicolumn{1}{c}{11} & \multicolumn{1}{c}{4(2/11/2/85)}\\
\multicolumn{1}{c}{86(86/13/0/1)} & \multicolumn{1}{c|}{1(0/0/1/99)} & \multicolumn{1}{c}{84} & \multicolumn{1}{c}{4(2/1/2/95)}\\
\multicolumn{1}{c}{2.15} & \multicolumn{1}{c|}{\textcolor{magenta}{28.13}} & \multicolumn{1}{c}{10.50} & \multicolumn{1}{c}{10.33}\\
\multicolumn{1}{c}{5.77} & \multicolumn{1}{c|}{5.15} & \multicolumn{1}{c}{1.19} & \multicolumn{1}{c}{3.54}\\
\multicolumn{1}{c}{0(0/90/0/10)} & \multicolumn{1}{c|}{3(0/0/3/97)} & \multicolumn{1}{c}{3} & \multicolumn{1}{c}{0(0/0/0/100)}\\
\multicolumn{1}{c}{47(44/45/3/8)} & \multicolumn{1}{c|}{49(0/0/49/51)} & \multicolumn{1}{c}{47} & \multicolumn{1}{c}{36(0/1/35/64)}\\
\multicolumn{1}{c}{62(41/13/21/25)} & \multicolumn{1}{c|}{52(0/0/52/48)} & \multicolumn{1}{c}{49} & \multicolumn{1}{c}{56(0/0/56/44)}\\
\multicolumn{1}{c}{34(18/35/16/31)} & \multicolumn{1}{c|}{37(0/0/37/63)} & \multicolumn{1}{c}{\textcolor{blue}{44}} & \multicolumn{1}{c}{12(0/0/12/88)}\\
\multicolumn{1}{c}{42(28/41/14/17)} & \multicolumn{1}{c|}{58(0/0/58/42)} & \multicolumn{1}{c}{56} & \multicolumn{1}{c}{40(0/0/40/60)}\\
\multicolumn{1}{c}{53(49/43/4/4)} & \multicolumn{1}{c|}{50(0/0/50/50)} & \multicolumn{1}{c}{56} & \multicolumn{1}{c}{45(0/0/45/55)}\\
\multicolumn{1}{c}{40(31/41/9/19)} & \multicolumn{1}{c|}{44(0/0/44/56)} & \multicolumn{1}{c}{\textcolor{blue}{49}} & \multicolumn{1}{c}{48(0/0/48/52)}\\
\multicolumn{1}{c}{65(58/29/7/6)} & \multicolumn{1}{c|}{52(0/0/52/48)} & \multicolumn{1}{c}{68} & \multicolumn{1}{c}{51(15/16/36/23)}\\
\multicolumn{1}{c}{\textbf{\textcolor{magenta}{0.7084}}} & \multicolumn{1}{c|}{\textbf{0.6412}} & \multicolumn{1}{c}{\textbf{\textcolor{blue}{0.7361}}} & \multicolumn{1}{c}{\textbf{0.4909}}\\
\bottomrule
\end{tabular}
\end{sc}
\end{small}
\end{center}
\vskip -0.1in
\end{table*}

\end{document}